\documentclass[
  a4paper,
  titlepage,
  twoside,
  parskip,
  bibliography=totoc,
  BCOR15mm
]{scrartcl}

\newcommand*{\sequence}[2]{\langle #1_1, ..., #1_#2 \rangle}

\usepackage[utf8]{inputenc}
\usepackage[T1]{fontenc}
\usepackage{lmodern}
\usepackage[german,english]{babel}
\usepackage{hyperref}
\usepackage{mathtools}
\usepackage[numbers]{natbib}
\usepackage{booktabs}
\usepackage{multirow}
\usepackage{placeins}

\usepackage{amsthm}
\usepackage{amssymb}
 
\theoremstyle{definition}
\newtheorem{definition}{Definition}[section]

\usepackage[ruled]{algorithm2e}
\usepackage{listings}

\usepackage{fancyhdr}
\fancypagestyle{empty-page}{
  \fancyhf{}
  \fancyfoot[LE,RO]{\thepage}
}
\providecommand{\tightlist}{%
  \setlength{\itemsep}{0pt}\setlength{\parskip}{0pt}}
  
\newcommand{\itemiz}{%
  \begin{itemize}
  \tightlist}

\begin{document}

  \pagenumbering{roman}
  \begin{titlepage}
  \titlehead{
    \centering
    \includegraphics[width=120pt]{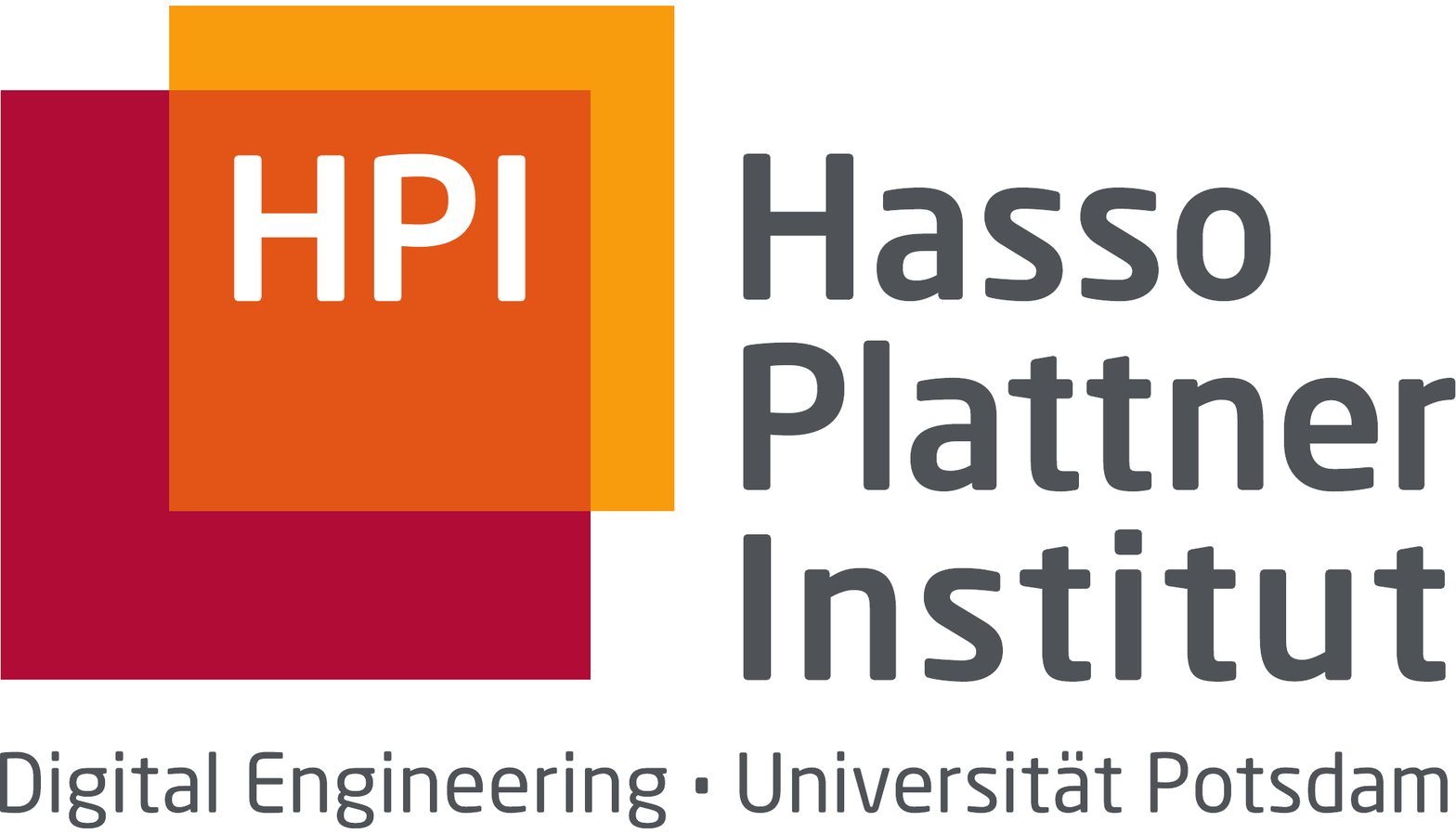}
  }
  \subject{Bachelor's Thesis}
  \title{Feature Learning for Meta-Paths in Knowledge Graphs}
  \subtitle{Feature Learning für Meta-Pfade in Wissensgraphen} 
  \author{Sebastian Bischoff\\{\small{\url{sebastian@salzreute.de}}}}
  \date{Submitted on July 30, 2018}
  \publishers{
    Knowledge Discovery and Data Mining Group\\
    Hasso Plattner Institute, Germany \par\vspace{2cm}
    Supervisors\\\smallskip Dr. Davide Mottin \\ Prof. Dr. Emmanuel Müller\vspace{-2cm}
  }
  \maketitle
  \end{titlepage}

  \cleardoublepage
  \section*{Abstract}

In this thesis, we study the problem of feature learning on heterogeneous knowledge graphs.
These features can be used to perform tasks such as link prediction, classification and clustering on graphs.
Knowledge graphs provide rich semantics encoded in the edge and node types.
Meta-paths consist of these types and abstract paths in the graph.

Until now, meta-paths can only be used as categorical features with high redundancy and are therefore unsuitable for machine learning models.
We propose meta-path embeddings to solve this problem by learning semantical and compact vector representations of them.
Current graph embedding methods only embed nodes and edge types and therefore miss semantics encoded in the combination of them.
Our method embeds meta-paths using the skipgram model with an extension to deal with the redundancy and high amount of meta-paths in big knowledge graphs.

We critically evaluate our embedding approach by predicting links on Wikidata.
The experiments indicate that we learn a sensible embedding of the meta-paths but can improve it further.
  \cleardoublepage
  \tableofcontents
  \cleardoublepage

  \pagenumbering{arabic}
  \section{Introduction}\label{sec:introduction}
The research community showed an increased interest in knowledge graphs over the last years. 
They are able to structure large amounts of information and can therefore be used in various domains and use cases.
One well-known and daily used application is in Google's search for entity disambiguation, enrichment of search results and exploratory search~\cite{nickel2016review}.
A knowledge graph could also be used as a source of knowledge for a future AI system.
How to provide general knowledge to such a system is one of the  problems of AI~\cite{lenat1992thresholds}.

Firstly, knowledge graphs or information networks are a very general way of storing information.
Therefore, methods that are based on them are generally applicable independent of the data or the domain.
Secondly, different fields use them for collecting information.
Biologists for example are building protein-protein interaction networks or gene regulatory networks to investigate the workings of proteins and genes.
Most of the data on the web is unstructured~\cite{blumberg2003problem}.
To make these resources usable, different approaches were proposed to extract knowledge graphs from unstructured data sources such as Wikipedia~\cite{suchanek2007yago} or the web.
Lastly, they can be used to combine information from multiple sources in the process of information fusion~\cite{shi2017survey}.

Such knowledge graphs can be used to perform tasks such as classification, prediction, clustering or anomaly detection.
The usefulness of these tasks depends on the data and the meaning in the domain.
The prediction of missing links, also called knowledge base completion, can help to suggest friends in a social network or to reduce the experimentation costs in biology.
Molecular interactions can be modeled as a link prediction problem on a corresponding graph.
Because 99.7\% of these interactions in human cells~\cite{stumpf2008estimating,amaral2008truer} are still unknown~\cite{lu2011link}, it would be helpful to focus the experiments on interactions with a high probability.
Another type of link prediction is the one where we want to predict future links in an evolving graph.
Concerning nodes, we can classify them in predefined classes, cluster them in groups with similar characteristics, detect anomalies~\cite{akoglu2015graph}, rank them for information retrieval~\cite{shi2017survey}, and recommend them to a user.

All current methods focus on embedding only nodes or nodes and edge types simultaneously.
To the best of our knowledge no other work proposes node type embeddings except \citet{xie2016representation} who learn multiple embeddings per type.
All the other publications work with triplets (head entity, relation, tail entity), not including node types.
Furthermore, there is no method for embedding meta-paths.
Meta-paths are sequences alternating node and edge types. 
They are an abstraction of paths in the underlying graph and therefore summarize many paths.

Methods for future link prediction based on meta-paths~\cite{sun2011co,yu2012citation} only use structural features and no semantic features.

However, knowledge graphs contain much richer information that is not used by the previously mentioned methods.
By representing specific concepts in the domain, meta-paths are capturing semantics which can not be represented when considering nodes and edges alone.
The correlation of meta-paths is more meaningful than the correlation of edges and nodes.
So one would expect a higher predictive power with meta-paths as features.
Furthermore, meta-paths span more than only the direct neighborhood and hence yield features with more information.
They extend over the direct neighborhood with the cost of a high time complexity when computing them.

The main problem is that meta-paths consist of categorical entities, namely node and edge types.
Almost all machine learning algorithms need vector representation of the input data.
The simplest way to transform meta-paths into vectors would be to encode them with an ID in the form of a one-hot representation.
The problem with this representation is that the number of meta-paths is exponential in the length of the meta-paths and the dimensionality grows linearly with the number of meta-paths.
Therefore, all machine learning algorithms will suffer from the curse of dimensionality.
Additionally, meta-paths are highly redundant because of the shared subparts and therefore this representation contains a lot of redundancy.

The need for vector representations of texts arose early in the data mining and information retrieval community.
They are needed for document classification, document retrieval, and document ranking.
The first approaches represented words by a combination of latent classes learned from an underlying text corpus~\cite{deerwester1990indexing,hofmann1999probabilistic,blei2003latent}.
Afterwards, the focus shifted to learning vector representations by optimizing a prediction based loss~\cite{mikolov2013efficient,mikolov2013distributed}.
The assumption in these embeddings is that words occurring together in a context share some form of meaning~\cite{harris1954distributional}.

We have to make some assumptions to use the methods developed in the text embedding community for meta-path embeddings.
We assume that meta-paths which occur in the same context also share some form of meaning.
The context is defined by all the meta-paths between two nodes.
If we take two actors that played together in a movie, played in different episodes of one TV series and went to the same (acting) school, then their nodes are connected by meta-paths representing these concepts.
All these concepts share at least the "dimension" \textit{acting} but maybe there are also other ones present such as \textit{occupation} and \textit{education}.

\pagebreak
\paragraph{Scope and limitations}
Prerequisite for our method is that edge and node types are defined on the knowledge graph.
There are knowledge graphs that do not satisfy this requirement but there are also many schema-based knowledge graphs~\cite{nickel2016review} such as Freebase~\cite{bollacker2008freebase}, Wikidata~\cite{vrandevcic2014wikidata}, DBpedia~\cite{auer2007dbpedia}, YAGO~\cite{suchanek2007yago,hoffart2013yago2} and Google Knowledge Graph that can be used~\cite{nickel2016review}.
They do not necessarily have to provide a node type directly.
It is sufficient that a class hierarchy is defined in the knowledge graph and that each node has an \textit{instance of}-relation with a class.
If meta-paths are given on a graph, we can compute embeddings for them in a reasonable time.
If meta-paths and their embeddings are only an intermediate step, one major drawback of our approach is that meta-paths are expensive to compute.
However, we do not need all meta-paths on the graph and we therefore propose a probabilistic mining algorithm with a reduced runtime.

\paragraph{Outcomes}
Our experiments show that the approach learns a sensible representation of meta-paths and especially their components, node and edge types. However, we have to refine the meta-path-based features for link prediction and perform additional experiments to evaluate our approach more thoroughly.

\paragraph{Contribution}
In this work, we propose meta-path embeddings targeting real-world knowledge graphs, node and edge type embeddings, an enhanced version of link prediction based on meta-paths and new vector representations for edges and nodes based on meta-paths.
Furthermore, we introduce an experiment stack to mine meta-paths, to transform them for the usage with text embedding implementations and to perform link prediction.

\paragraph{Bachelor Project}
In our bachelor project, we built an interactive exploration tool for knowledge graphs~\cite{behrens2018metaexp} motivated by a biological use case of our project partner.
The exploration tool incorporates the domain knowledge of a user into the exploration process by querying ratings of meta-paths from the user.
As mentioned earlier, the number of meta-paths is very high and exceeds the number of ratings a user wants to provide to the system.
To overcome this problem, the system uses active learning to only query ratings for a selection of meta-paths~\cite{Ziegler2018Active}.
This selection is based on how much new information the meta-paths provide.
Simultaneously, it learns a predictor for the user rating and uses it for all meta-paths the user did not rate.
This part requires manual feature engineering or an embedding of the meta-paths.
Lastly, the naive computation of meta-paths turned out to be infeasible.

\paragraph{Structure of the Thesis}
In Section \ref{sec:related-work}, we discuss the current body of work related to our approach, in Section \ref{sec:feature},  we introduce our meta-path embedding model, node and edge embeddings based on the model and a meta-path mining algorithm, in Section \ref{sec:evaluation}, we present different experiments to evaluate our approach and compare it with other approaches, in Section \ref{sec:implementation}, we discuss some implementation details, and conclude our work in Section \ref{sec:conclusion}.

  \section{Related Work}\label{sec:related-work}
The body of work related to our method can be divided in the group of graph embedding on classical graphs and knowledge graphs, text embedding, meta-paths-based works and link prediction.
In the following, we briefly summarize the different fields and compare the approaches.

\paragraph{Meta-paths}
Meta-paths~\cite{sun2011pathsim,meng2015discovering} were first introduced for the definition of Pathsim~\cite{sun2011pathsim}, a similarity measure on heterogeneous graphs.
Later, HeteSim~\cite{shi2012relevance,shi2014hetesim} extended Pathsim by allowing to measure the similarity of objects with different types and removing the restriction on symmetric paths.
Meta-paths are further used in tasks like user-guided entity clustering~\cite{sun2013pathselclus}, link prediction\cite{yu2012citation,sun2012will,sun2011co}, multi-network collective link prediction~\cite{zhang2014meta}, collective classification~\cite{kong2012meta}, entity similarity search~\cite{yu2012user}, and entity ranking~\cite{liu2014meta}.

\paragraph{Node embedding}
Methods for node embedding like node2vec~\cite{grover2016node2vec}, DeepWalk~\cite{perozzi2014deepwalk}, LINE~\cite{tang2015line}, VERSE~\cite{tsitsulin2018verse}, HOPE~\cite{ou2016asymmetric} and GraRep~\cite{cao2015grarep} embed the structure of a graph without edge and node types.
Graph factorization models~\cite{ahmed2013distributed} use matrix factorization for node embedding.

DeepWalk~\cite{perozzi2014deepwalk} performs random walks on the graph, it treats walks as sentences and embeds them using the word2vec skipgram model~\cite{mikolov2013efficient,mikolov2013distributed}.
Similarly, we mine the meta-paths in a random walk fashion and treat meta-paths as words because we embed using the principle that meta-paths in the same context have a similar meaning.
Whereas, DeepWalk uses the principle that nodes in the same context have a similar meaning.

\paragraph{Knowledge graph embedding}
When we work with knowledge graphs, we have more information available to embed nodes, edge types and node types.
The first group of methods performs node embedding by incorporating the extra information from knowledge graphs.
The neighborhood mixture model~\cite{nguyen2016neighborhood} uses the embedding of neighboring nodes and the connecting edges to calculate a node embedding.
It is based on methods which produce node and edge embeddings like TransE~\cite{Bordes2013} and TransH~\cite{Wang2014}.
Metapath2vec~\cite{dong2017metapath2vec} embeds after the same principle as node2vec~\cite{grover2016node2vec}, DeepWalk~\cite{perozzi2014deepwalk} and LINE~\cite{tang2015line} does, using the skipgram model. 
They extend the node embedding methods on simple graphs by proposing a way to incorporate the information in heterogeneous graphs.
To achieve this, they condition random walks, producing "sentences", on user-specified meta-paths.
One problem with this approach is that the results strongly depend on the selected meta-path(s) and that the user needs a deeper understanding of the graph to choose them correctly. 
They referred in their experiments to other publications~\cite{huang2016meta,sun2012mining,sun2011pathsim,sun2013pathselclus} to find the best working meta-paths for tasks on academic networks.
This shows how their experiments specifically target the data sets and that a more general meta-path-based approach would be favorable.
\citet{luo2015context} treat meta-path instances which they call \textit{knowledge paths} as words and embed them with the skip-gram model.
In a second step, they refine the embedding with SME~\cite{bordes2014semantic}, TransE~\cite{Bordes2013} or SE~\cite{bordes2011learning}.

\vspace{-5pt}
Another body of work learns node and edge embeddings jointly.
\citet{Garcia-Duran2015} groups them in two-way approaches and three-way approaches.
Two-way approaches such as SME~\cite{bordes2014semantic}, TransE~\cite{Bordes2013}, TransH~\cite{Wang2014} and TransR~\cite{Lin2017} model the interactions between the head and the tail, the head and the label, and the label and the tail of knowledge graph triplets.
Three-way~\cite{Garcia-Duran2015} approaches such as TATEC~\cite{Garcia-Duran2015} jointly model the interaction of head, label and tail.
Two-way approaches use vectors and three-way approaches use matrices to model the relation.
All these approaches mainly differ in the parameterization of nodes and edges as vectors or matrices and their cost function~\cite{wang2017knowledge}.
These works learn only from the knowledge triplets and therefore do not capture the structure of the graph.
Furthermore, they do not explicitly model the node types and can only learn from them implicitly, if a taxonomy is defined in the triplets they use to learn their embeddings.
It remains an open question how well the node types are captured.
Another line of work uses multi-layer fully connected neural networks~\cite{nickel2016holographic} and convolutional neural networks such as ConvE~\cite{dettmers2017convolutional} and ConvKB~\cite{nguyen2018novel} to map triplets to an embedding.

\vspace{-5pt}
Only \citet{xie2016representation} and \citet{guo2015semantically} use node types in their graph embedding.
\citet{xie2016representation} learns a projection matrix for each node type, where a node type has a specific embedding dependent on the relation it occurs in.
\citet{guo2015semantically} proposes a semantically smooth embedding where nodes with the same node types should be closely located to each other in the embedding space~\cite{wang2017knowledge}.

\vspace{-5pt}
\begin{table}
\centering
\begin{tabular}{*{5}{c}}
\toprule
\textit{subset} & Nodes & Node types & Edge types & Meta-paths \\
\midrule
Node embedding techniques & \multirow{2}{*}{x} & & &  \\
\cite{grover2016node2vec,perozzi2014deepwalk,tsitsulin2018verse,tang2015line,ou2016asymmetric,cao2015grarep} & & & & \\
Translational approaches & \multirow{2}{*}{x} &  &\multirow{2}{*}{x} &  \\
\cite{Bordes2013,Wang2014,bordes2014semantic,bordes2011learning,Lin2017,lin2015modeling}  &  & & & \\
TKRL~\cite{xie2016representation} & x & (x) & x & \\
Neighboorhood mixture & \multirow{2}{*}{x} & & & \\
model~\cite{nguyen2016neighborhood}  &  & & & \\
metapath2vec~\cite{dong2017metapath2vec}  & x & & & \\
Our method  & x & x & x & x\\
\bottomrule
\end{tabular}
\caption{Comparison of different graph embedding methods regarding the objects they embed.}
\label{tab:related}
\end{table}
Another line of work uses relation paths which are like meta-paths but without node types.
They only include edge types.
Relation path embedding such as PTransE~\cite{lin2015modeling} tries to embed nodes and relationships in a knowledge graph so that the combination of relationships is a sensible embedding for the path composed of this relationships.
\citet{toutanova2016compositional} include nodes into the relation paths.
In comparison, node types in the relation path (= meta-path) help us to generalize because two paths with different nodes but the same node types are learned independently by \citet{toutanova2016compositional} and jointly in our method. 
Path queries~\cite{guu2015traversing} such as \textit{What parts of an Airbus A380 are made out of steel?} that would translate to \textit{A380 - has\_part - part - material\_used - steel} can be interpreted as relation paths~\cite{nguyen2017overview} and if the user would specify node types in their query such as \textit{part} also as meta-paths.
Consequently, we can use meta-paths to enhance the answering of path queries.

Lastly, there are many other works using additional data such as text with mentions of knowledge graph entities~\cite{toutanova2015observed} for the embedding which improves the performance~\cite{toutanova2016compositional} or using other features from the graph such as subgraphs generated with random walks for knowledge base completion~\cite{wang2016knowledge}.
All these approaches are compared in Table \ref{tab:related}.

\vspace{-5pt}
\paragraph{Text embedding}
The first text embedding approaches based on latent class models were methods such as latent semantic analysis~\cite{deerwester1990indexing}, probabilistic latent semantic indexing~\cite{hofmann1999probabilistic} and latent dirichlet allocation~\cite{blei2003latent} in the field of distributional semantics~\cite{turney2010frequency,baroni2010distributional}.
The distributional hypothesis~\cite{harris1954distributional} suggests that words in the same context often have a similar meaning.
Based on this hypothesis, predictive models were first proposed for embedding words~\cite{mikolov2013efficient,mikolov2013distributed} and later also for  sentences~\cite{le2014distributed,gupta2016doc2sent2vec}, paragraphs~\cite{le2014distributed}, and documents~\cite{gupta2016doc2sent2vec}.
Another group of approaches uses the matrix factorization of the word co-occurrence matrix~\cite{pennington2014glove,murphy2012learning,levy2014neural,salle2016matrix}.

The embedding of sentences and paragraphs with paragraph vectors~\cite{le2014distributed} could be transferred to meta-paths by treating one meta-path as a sentence.
This would result in a model which does not take the high redundancy of meta-paths into account and would learn inefficiently.

An addition to the skipgram model~\cite{bojanowski2017enriching} uses character n-grams for each word to capture subword information.
A word is represented as the sum of the embeddings of n-grams contained in the word.

Even smaller entities are considered in character level language models trained with recurrent neural networks~\cite{mikolov2012subword,sutskever2011generating,graves2013generating} or convolutional neural networks~\cite{kim2016character}.
Other ways to calculate features for texts are based on n-grams or are built by hand with features which are believed to perform well on the specific task.

\paragraph{Link prediction}
Different tasks using these embeddings were proposed.
The task of predicting missing links, named knowledge base completion, got the most attention.
All models producing embeddings for nodes such as Unstructured~\cite{bordes2014semantic} or nodes and edges such as TransE~\cite{Bordes2013} can be used for a scoring function $f(h,t)$ or $f(h,r,t)$ of the plausibility of a knowledge base triplet $(h,r,t)$~\cite{nguyen2017overview}.
Another task defined on these triplets is triplet classification~\cite{socher2013reasoning} where a classifier predicts if a given triplet exists in the knowledge graph.
The link prediction over evolving graphs was first introduced in the social sciences~\cite{liben2007link}.
It is a harder link prediction problem than in knowledge graph completion because a classifier has to learn the formation mechanisms of links~\cite{davis2011multi} instead of modeling the likelihood of links in a given graph.

Special tasks on academic networks such as DBLP\footnote{\url{https://dblp.uni-trier.de}} are citation and co-authorship prediction.
\citet{sun2011co} approach co-authorship prediction by searching all meta-paths between authors, calculate structural values for all these meta-paths and then learn a classifier to predict, if two authors will write a paper together in the future.
This approach only works for graphs with a small number of node and edge types because otherwise the number of dimensions of the feature space spanned by the meta-paths is very high.
Our method does not suffer from this problem because it reduces the number of dimensions.
Furthermore, we are not only considering the topology of the graph but also the semantics of the meta-paths.
\citet{yu2012citation} propose a method for citation prediction also based on structural features of meta-paths such as the number of instances.
They do not use any semantic features of meta-paths.

An extended version of this problem is the estimation of the point in time a link will occur. 
\citet{sun2012will} mine meta-paths between the two nodes of an edge and calculate different measures based on them as features for a classifier.
We also use the meta-paths between two nodes in our meta-path-based edge embedding in Section \ref{sec:edges-mps} but we are using the combined embeddings of the meta-paths, instead of the measures of a meta-path, as features for a relationship building time predictor.
We could incorporate the topological features of \citet{sun2012will,sun2011co} and \citet{yu2012citation} as a weighting of the embeddings of the single meta-paths.

The path ranking algorithm~\cite{lao2010relational,gardner2015efficient,wang2016knowledge} uses relation paths as features for tasks like entity ranking.
This could be extended by using meta-paths instead of relation paths.
  \section{Meta-path Embedding}\label{sec:feature}
Many different topological features such as common neighbors, 
preferential attachment~\cite{barabasi2002evolution,newman2001clustering}, $Katz_\beta$~\cite{katz1953new}, $Adamic/Adar$~\cite{adamic2003friends}, and $PropFlow$~\cite{lichtenwalter2010new} were proposed for homogeneous networks.
Most of them are not usable for heterogeneous networks~\cite{sun2012will}.
Instead, we can use meta-paths as topological features in heterogeneous networks~\cite{sun2011co} because meta-paths encode topological features~\cite{sun2012will}.
Meta-paths are a very general representation which can even be transferred between graphs, if a mapping of the node and edge types is defined.
Meta-paths alone are not sufficient for the definition of these features.
We have to use measures such as instance counts and random walks~\cite{sun2011co} to quantify a meta-path. 

A predictor trained with these features lacks any information about the "meaning" of a meta-path because the previous features only concern the structure of a graph.
If we would have features describing the semantics of a meta-path, we could combine them with the structural features to get a holistic feature set.

To get these semantical features and overcome the representation of meta-paths as categorical features, we propose meta-path embeddings.
By embedding meta-paths, we get a compact representation of the enormous amount of (meta-)paths in real world knowledge graphs.
Additionally, we reduce the redundancy in the meta-paths and transform them to a representation machine learning algorithms can easily use.

To derive a meta-path embedding from text embedding approaches, we map the different parts of a word embedding to a graph as follows
\itemiz
\item The text corpus equals the whole graph.
\item One sentence equals all meta-paths between two nodes. 
Thereby, defining the context of one meta-path. 
This definition comes from the assumption that meta-paths which occur in the same context also share some form of meaning.
The underlying intuition is that meta-paths which connect the same nodes in a graph are similar.
\item One word equals one meta-path.
\end{itemize}

We deal with the redundancy of meta-paths by using ideas for capturing subword information.
\citet{bojanowski2017enriching} propose an extension of the skip-gram model~\cite{mikolov2013distributed,mikolov2013efficient} by including character-level n-grams into the embedding.
This results in an embedding which takes advantage of shared parts in word families.
In our case, the parts of a word equal parts of a meta-paths.
These parts are combinations of node and edge types which also occur in different meta-paths.

\pagebreak
The redundancy of meta-paths is an observed problem which is addressed by \citet{kong2012meta} with the decomposition of meta-paths in smallest, non-trivial meta-paths.
They argue that many, and in part redundant, meta-paths are noise for a classifier.
We deal with the redundancy by  capturing the subword information and projecting to the embedding space.

In the following parts, we first introduce the necessary definitions for this work, our embedding model, different features for nodes and edges and describe the mining of meta-paths for the embedding.

\subsection{Background}\label{sec:background}
In this chapter, we introduce the definition of knowledge graphs, their homogeneity and heterogeneity, meta-paths on the network schema and embeddings.

\theoremstyle{definition}
\begin{definition}[Knowledge graph~\cite{sun2013mining,sun2009ranking}]
A knowledge graph, also named information network, is a directed graph $G=(V,E,\varphi,\psi)$ with a set of node types $\mathcal{A}$ and edge types $\mathcal{R}$. 
A node type mapping function $\varphi: V\mapsto \mathcal{A}$ maps nodes to their node types and an edge type mapping function $\psi: E\mapsto \mathcal{R}$ maps edges to their edge types.
Furthermore, the edge types imply the node types as follows $\psi(e_1=(v_1,u_1)) = \psi(e_2=(v_2,u_2)) \implies \varphi(v_1)=\varphi(v_2) \wedge \varphi(u_1)=\varphi(u_2)$~\cite{shi2017survey}.
\end{definition}

\begin{definition}[Homogeneous/heterogenous knowledge graph]
If a knowledge graph has multiple node types $|\mathcal{A}| > 1$ or edge types $|\mathcal{R}| > 1$, we call it a heterogeneous knowledge graph or heterogeneous information network (HIN).
We call a knowledge graph a homogeneous knowledge graph, if it has only one node type $|\mathcal{A}| = 1$ and edge type $|\mathcal{R}| = 1$.
\end{definition}

\begin{definition}[Network schema~\cite{sun2013mining,sun2009ranking}]
$S_G = (\mathcal{A},\mathcal{R})$ is a network schema of a knowledge graph $G=(V,E,\varphi,\psi)$.
\end{definition}

\begin{definition}[Meta-path~\cite{sun2011co,sun2011pathsim}]
The meta-path associated with the path $\sequence{n}{t}, n_i \in V, 1 \leq i \leq t$ is a path on the network schema $S_G = (\mathcal{A},\mathcal{R})$ with length $t$.
It is a sequence $\mathcal{P}: \langle \varphi(n_1),\psi((n_1, n_2)), ... ,\psi((n_{t-1}, n_t)), \varphi(n_t)\rangle$ that alternates node and edge types along the path~\cite{behrens2018metaexp}.

A meta-path instance is a path in the knowledge graph where the nodes and edges have the types specified in the meta-path.
\end{definition}

\begin{definition}[Embedding]
An embedding $f:k \to r$ with embedding size $n$ maps  $f: K \mapsto \mathbb{R}^n$.
A meta-path embedding $emb: mp \to r$ therefore maps $emb:\mathcal{MP} \mapsto \mathbb{R}^n$.
%
%
\end{definition}

\subsection{Embedding Model\label{sec:embedding}}
Our model and formalization follows \citet{bojanowski2017enriching} but is adapted to meta-paths.
We have a meta-path vocabulary $MP$ of size $|MP|$ and denote each meta-path by its index $mp \in {1,...,|MP|}$.
Our goal is to learn a vector representation for each $mp$.
The skipgram model needs a training corpus of size $T$ denoted as $mp_1,...mp_T$.
The training objective of the skipgram model is to maximize the probability of the meta-paths in the context given the meta-path given by
\begin{equation*}
\label{eq:skipgram}
\sum_{t=1}^{T}{\sum_{c\in C_t}^{}{log\ p(mp_c|mp_t)}}
\end{equation*}
$C_t$ being the context of the meta-path with index $t$ and $c$ is one meta-path in the context.
The context is defined as a subset of all meta-paths between two nodes.
With $m$ meta-paths between two nodes and the equivalent of a sentence being $n$ meta-paths long, we have $\binom{m}{n}$ possible sentences because the graph does not define a order of the meta-paths.
In real world knowledge graphs, $m$ is rather high and the therefore resulting high number of sentences causes a need to sample from them.

We then express the prediction of the meta-paths in the context $C$ as binary classification tasks with the logistic loss function $l: x \mapsto log(1+e^{-x})$ and a set $n \in \mathcal{N}_{t,c}$ of negative examples drawn randomly from the the vocabulary $MP$ which results in the objective

\begin{equation*}
\label{eq:objective}
\sum_{t=1}^{T}{\Big[\sum_{c\in C_t}^{}{l(s(mp_t,mp_c))+\sum_{n \in \mathcal{N}_{t,c}}^{}{l(-s(mp_t,n))}}\Big]}
\end{equation*}

Now, it only remains to define the scoring function $s$.
We use the embedding vectors $v_{mp_t}$ of the currently selected meta-path and the one of the meta-path in the context $v_{mp_c}$ and define $s(mp_t,mp_c)={v_{mp_t}}^\top v_{mp_c}$.

We could embed meta-paths only with the skipgram model but this would not address the redundancy of them.
For this reason, subword information is added to the skipgram model.
The subword information is captured by an embedding of n-grams.
In the extended model, we additionally have a dictionary of n-grams $G$ with $min\leq n\leq max$ where $min$ is the minimal length and $max$ the maximal length of n-grams in $G$.
The n-grams in meta-path $mp$ are denoted as $\mathbf{G_{mp}}$.
We define the scoring function $s(mp,c)$ in the subword model as 
\begin{equation*}
\label{eq:subword}
s(mp,c) = emb(mp)^\top emb(c)
\end{equation*}

Furthermore, the embedding of one meta-path is defined as
\begin{equation*}
emb(mp) = \sum_{\mathbf{g \in G_mp}}^{}{emb(g)}
\end{equation*}
This also produces node type $emb_{nodetype}(i)$ and edge type $emb_{edgetype}((i,j))$ embeddings, if we choose $min = 1$. 
We would not expect a very good performance in comparison with future methods specifically designed for node type embeddings.
These embeddings are the baseline version of our method.
%
%
Based on this model, a definition of different node and edge representations is possible.

\subsubsection{Meta-path-based Edge Embedding}\label{sec:edges-mps}
An edge $u=(i,j)$ can be represented by all the meta-paths connecting nodes $i$ and $j$.
Given $i,j\in V$, $\mathcal{MP}(i,j)$ is defined as all meta-paths between nodes $i$ and $j$ and $emb_{edge}: E \mapsto \mathbb{R}^n$ as edge embedding function.
The representation of the edge $(i,j)$ is defined as
\begin{equation*}
\label{eq:edges-mps}
emb_{edge}((i,j))=\frac{1}{|\mathcal{MP}(i,j)|}\sum_{mp\in \mathcal{MP}(i,j)}{emb(mp)}
\end{equation*}

\subsubsection{Meta-path-based Node Embedding}\label{sec:nodes-mps}
Given $i\in V$, $\mathcal{MP}(i)$ is defined as all meta-paths starting or ending in node $i$ and $emb_{node:mps}: V \mapsto \mathbb{R}^n$ as node embedding function.
The representation of node $i$ is defined as
\begin{equation*}
\label{eq:nodes-mps}
emb_{node: mps}(i)=\frac{1}{|\mathcal{MP}(i)|}\sum_{mp\in \mathcal{MP}(i)}{emb(mp)}
\end{equation*}

\subsubsection{Node type-based Node Embedding}\label{sec:nodes-types}
We extend $\varphi$ to $\varphi'(i)$, $\varphi'$ returns the set of node types of node $i$.
Given $i\in V$ and $j\in \varphi'(i)$, $emb_{node type}(j)$ is defined as embedding of node type $j$.
The embedding of the node types is defined as $emb_{node type}: \mathcal{A} \mapsto \mathbb{R}^n$ and the embedding of one node based on its node types is defined as $emb_{node:node types}: V \mapsto \mathbb{R}^n$.
The representation of node $i$ is defined as
\begin{equation*}
\label{eq:nodes-types}
emb_{node: node types}(i)=\frac{1}{|\varphi(i)|}\sum_{u\in \varphi(i)}{emb_{node type}(u)}
\end{equation*}
One could further add a parameter to this definition weighting the depth of the node type in the class hierarchy.
With this parameter, one could give more weight to fine-grained classes because they cover the details of a node.

\citet{liben2007link} state that it is not sensible to predict edges where one of the nodes is not present in the training data.
In principle, we can produce embeddings for any nodes, if the node types of the new nodes are provided and present when training the embedding.

We could further define an individual node embedding by first embedding all nodes with their node types and then learning a delta to represent the specific characteristics of each node.
Given $u \in V$, $\sum_{w \in \varphi'(u)}{emb_{node type}(w)}$ is defined as the embedding of the node types $\varphi'(u)$ of $u$ and  $emb_{\Delta}(u)$ as the embedding of only node $u$. $emb_{node:delta}(u)$ as the embedding of $u$ is defined as

\begin{equation*}
emb_{node:delta}(u)=\sum_{w \in \varphi'(u)}{emb_{node type}(w)}+emb_{\Delta}(u)
\end{equation*}

\subsection{Meta-path Mining}
If no meta-paths are given, we have to calculate them.
In the following, we introduce our algorithm, demonstrate its problems and propose an improved version.

Generally, the graph is treated as undirected and the following algorithms assume that each node has only one node label.
The algorithm can easily be extended to multiple types per node by collecting the types along the meta-path and forming the cartesian product in the end.

\FloatBarrier
The parameters for Algorithm \ref{alg:MPMining} are a knowledge graph $KG=(V,E,\varphi,\psi)$ consisting of a graph $G=(V,E)$  with node labeling function $\varphi$ and edge labeling function $\psi$ and $metaPathLength$ specifying the maximum meta-path length.
The algorithm works in a breadth-first manner and calls the function \textit{computeMetaPaths} for each node in the graph.

Function \textit{computeMetaPaths} additionally gets the $metapath$ up to this point, start node $u$ and current node $v$ as input.
It saves the meta-path which is extended by the label of the current node as a meta-path between the start node and the current node.
Afterwards, it recursively calls itself for each edge of the current node after adding the edge label to the meta-path.

\begin{algorithm}
\label{alg:MPMining}
\SetKwData{metapath}{metapath}\SetKwData{This}{this}\SetKwData{Up}{up}\SetKwData{AdjacentNodes}{adjacentNodes}
\SetKwFunction{computeMetaPaths}{computeMetaPaths}\SetKwData{metapaths}{metapaths}\SetKwFunction{Label}{Label}
\SetKwInOut{Input}{input}\SetKwInOut{Output}{output}
\SetKwProg{Fn}{Function}{}{}

 \Input{$KG=(V,E,\varphi,\psi)$, $metaPathLength$}
 \Output{A dictionary $metapaths$ containing for $metapaths$[u, v] all meta-paths between nodes $u$ and $v$}
 \BlankLine
 
 \Fn{computeMetaPaths(int $metaPathLength$)}{
 \For{$v \in V$}{
  \computeMetaPaths{$v$, $v$, $metaPathLength$, empty \metapath}\;
 }
 }
 \BlankLine
 \BlankLine
 
 \Fn{computeMetaPaths(current node $v$, start node $u$, int $metaPathLength$, Meta-path $metapath$)}{
   \If{$metaPathLength <= 0$}{
  	return\;
  }
  \metapath.append($\varphi(v)$)\;
  \metapaths[$u$, $v$].append(\metapath)\;
  \AdjacentNodes$\leftarrow \{x|(w,x)\in E$ or $(x,w)\in E\}$\;
  \For{$y \in \AdjacentNodes$}{
  	\metapath.append($\psi(v, y)$)\;
  	\computeMetaPaths{$y$, $u$, $metaPathLength - 1$, \metapath}
  }
}

\caption{Meta-Path Mining}

\end{algorithm}
\pagebreak
The worst case runtime of this algorithm is $\mathcal{O}(|V||V|^{length})=\mathcal{O}(|V|^{length+1})$ for a fully connected graph.
Luckily, knowledge graphs are in most cases not fully connected and the average case runtime is a more realistic approximation.
The average case runtime is $\mathcal{O}(|V|*\widetilde{deg}^{length})$ with the average node degree $\widetilde{deg}$, if we assume that the degree is equally distributed over all nodes.
This makes a big difference in knowledge graphs such as Wikidata because the average degree $\widetilde{deg} \approx 5 \ll |V|\approx 4*10^7$.

The memory consumption of our algorithm is mainly determined by the number of meta-paths.
The worst case number of meta-paths is $\mathcal{O}((|\mathcal{A}||\mathcal{R}|)^{length-1}|\mathcal{A}|)$, if the schema of the graph is fully connected.
The average case number of meta-paths is roughly $\mathcal{O}((avg.nodetypes*avg.degree)^{length-1}*avg.nodetypes)$ with $avg.nodetypes$ being the average number of node types per node, if we assume that $avg.nodetypes$ is  equally distributed over all nodes.
$avg.degree$ is hereby the upper bound of the number of edge types for the edges connected with one node.
The number of meta-paths scales with the number of node and edge types, which results in a high number of meta-paths in knowledge graphs.

\pagebreak
As one can see, the algorithm has a high runtime complexity for calculating a huge number of meta-paths.
One can compare our situation with the attempt to collect all text documents written at any time to accumulate all possible contexts of one word.
With this comparison, it is clear that the complete enumeration of meta-paths is not needed for the embedding.

Following this finding, we can mine meta-paths probabilistically by skipping nodes and edges in the traversal.
The result is equally to first calculating all meta-paths and then sampling from them with the exception that longer meta-paths are more rarely found by our algorithm.
The reasons for this is that the probability of one specific path is $(1-edgeSkipProbability)^{length}$ and therefore gets smaller if we consider longer paths.
\FloatBarrier
Algorithm \ref{alg:probMPMining} gets a knowledge graph $KG=(V,E,\varphi,\psi)$ consisting of a graph $G=(V,E)$  with node labeling function $\varphi$ and edge labeling function $\psi$ and $metaPathLength$ specifying the maximum meta-path length as input like the deterministic Algorithm \ref{alg:MPMining}.
Additional inputs are the $nodeSkipProbability$ specifying the probability of skipping one node in the outer loop and the $edgeSkipProbability$ specifying the probability of skipping one edge in the recursion.
The addition in the probabilistic algorithm is that it skips nodes when it calls the function \textit{probabilisticallyComputeMetaPaths} for each node in the graph.

The function \textit{probabilisticallyComputeMetaPaths}  gets the $metapath$ up to this point, the start node $u$, the current node $v$ and $edgeSkipProbability$ specifying the probability of skipping one edge as input.
It only recursively calls itself if a random draw was successful and skips the edge otherwise.
This algorithm produces the same results as performing random walks with a restart probability of $edgeSkipProbability$, a restart probability of 1 if the random walk reaches length $length$ and a uniformly distributed transition probability would.
But due to the possible repeated traversal of the same edge, random walks are inefficient compared to our algorithm.

\begin{algorithm}
\SetKwData{AdjacentNodes}{adjacentNodes}\SetKwData{metapath}{metapath}\SetKwData{nodeSkipProbability}{nodeSkipProbability}\SetKwData{edgeSkipProbability}{edgeSkipProbability}\SetKwData{metapaths}{metapaths}
\SetKwFunction{probabilisticallyComputeMetaPaths}{probabilisticallyComputeMetaPaths}\SetKwFunction{random}{random}
\SetKwInOut{Input}{input}\SetKwInOut{Output}{output}
\SetKwProg{Fn}{Function}{}{}
 
 \Input{$KG=(V,E,\varphi,\psi)$, $metaPathLength$, $nodeSkipProbability$, $edgeSkipProbability$}
 \Output{A dictionary $metapaths$ containing for $metapaths$[u, v] meta-paths between nodes $u$ and $v$}
  \BlankLine
 
 \Fn{probabilisticallyComputeMetaPaths(int $metaPathLength$, float $edgeSkipProbability$, float $nodeSkipProbability$)}{
 \For{$v \in V$}{
  \If{\random{} > \nodeSkipProbability}{
   \probabilisticallyComputeMetaPaths{$v$, $v$, $metaPathLength$, empty \metapath, \edgeSkipProbability}\;
  }
 }
 }
 \BlankLine
 \BlankLine
 
  \Fn{probabilisticallyComputeMetaPaths(current node $v$, start node $u$, int $metaPathLength$, Meta-path $metapath$, float $edgeSkipProbability$)}{
   \If{$metaPathLength <= 0$}{
  	return\;
  }
  \metapath.append($\varphi(v)$)\;
  \metapaths[$u$, $v$].append(\metapath)\;
  \AdjacentNodes$\leftarrow \{x|(w,x)\in E$ or $(x,w)\in E\}$\;
  \For{$y \in \AdjacentNodes$}{
   \If{\random{} > \edgeSkipProbability}{
  	\metapath.append($\psi(v, y)$)\;
  	\computeMetaPaths{$y$, $u$, $metaPathLength - 1$, \metapath, \edgeSkipProbability}
   }
  }
  }
  
 \caption{Probabilistic Meta-Path Mining}
\label{alg:probMPMining}
\end{algorithm}

The embedding model for the modified algorithm does not change with the exceptions that the meta-path vocabulary $MP$ gets $\widetilde{MP}_{p,q}$ of size $|\widetilde{MP}_{p,q}|$ and the context $C_t$ of the meta-path with index $t$ gets $\tilde{C}_{t,p,q}$ with node skip probability $p$ and edge skip probability of $q$.
We assume that the Meta-path vocabulary $\widetilde{MP}$ is roughly the same as $MP$ because one meta-path occurs many times. 
Therefore, it is unlikely that we miss it completely.

\pagebreak
\citet{neelakantan2015compositional} and \citet{lin2015modeling} select a subset of relationship paths through sampling and pruning which is similar to the probabilistic meta-path mining.
When executing our algorithm on a graph with a high number of node types, a higher $edgeSkipProbability$ and lower $nodeSkipProbability$ is preferable to start the mining from all node types even if it does not discover the whole neighborhood.

By not subsampling the graph in the beginning, we achieve an embedding of all node and edge types.
Skipping edges and nodes results in a reduced but still big enough amount of training data.
If we would first subsample the graph, we could calculate the full amount of training data for the embedding.
However, this embedding would not contain all node and edge types and would therefore be only limited in its applicability.
  \section{Evaluation}\label{sec:evaluation}
In this section, we evaluate our embedding model by comparing it with baseline methods and showing the influence of different parameters.

We can not use standard datasets such as Freebase15k~\cite{bordes2013translating} and Freebase1M~\cite{bordes2013translating} (\textit{not published}) to compare our method to previously published results.
These datasets only include edge types and lack node types.
This would not be a problem, if the edges would define a taxonomy.
Unfortunately, the \textit{type/object/type} relation~\cite{bast2014easy} which defines the taxonomy of Freebase is not included in Freebase15k.
Without the taxonomy, we can not define node types.

The other kind of dataset used in previous works are academic knowledge graphs such as DBLP.
Generally speaking, they have a very small schema and for that reason only very few meta-paths.
Our method is designed for rich schemata and we can not evaluate it meaningful with only very few node and edge types.

\paragraph{Dataset}
We use Wikidata~\cite{vrandevcic2014wikidata} for our experiments because it is a constantly growing, freely available and large knowledge graph.
Wikidata consists of entities which are either items or properties.
Properties are used to link items together.
Items are "all the things in human knowledge, including topics, concepts, and objects"\footnote{\url{https://www.wikidata.org/wiki/Help:Items}}.
Abstract concepts can be linked together with properties such as \textit{part of}, \textit{subclass of} and \textit{is instance of} and objects with properties such as \textit{material used}, \textit{brand} and \textit{has part}.
Claims are statements about a feature of an item which consists of a property and a value.
Wikidata can be transformed into a graph defining the items, properties and claims as nodes and the links between the nodes defined with properties as edges.

The Wikidata snapshot from 19th March 2018 used in the following experiments has roughly 45 770 000 entities and 237 950 000 claims.
Claims do not carry useful information for embedding meta-paths because their values are for example coordinates, geographic shapes, quantities and URLs.
Hence, we can exclude claims from our experiments.

There are no explicit node types in Wikidata, only a class hierarchy itself expressed as nodes linked together with \textit{subclass of} properties.
We use this class hierarchy as possible node types in the graph.
We follow the $u$ - \textit{is instance of} - $v$ relation of each node $u$, which is not in the class hierarchy, to assign the node label of node $v$ to it.
The class hierarchy forms a tree of height 40 and is very fine grained in the lower parts.
We reduce the tree to a height of $n$ by traversing it from the root and adding the class labels of the higher nodes to the lower ones up to depth $n$.
All levels from there to the leaves get the class labels from above.
This causes an abstraction of the fine-grained classes and enables us to learn an embedding of each class and its superclasses simultaneously.

\subsection{Link Prediction in Knowledge Graphs}\label{sec:link_prediction}
We use the problem of link prediction to evaluate the quality of our embedding.
In the following, we evaluate two of our three proposed feature definitions for nodes and edges from Sections \ref{sec:edges-mps}, \ref{sec:nodes-mps} and \ref{sec:nodes-types}.

We use the Wikidata version from the  19th March 2018 (time step $t_0$) and 2nd April 2018 (time step $t_1$) for our experiments and transform them as described above.
We set the height of the class hierarchy to 3 and mine meta-paths of $length$ 5 with a $nodeSkipProbability$ of 0.9999 and an $edgeSkipProbability$ of 0.99 to get a reasonable amount of data.
An incomplete run with $nodeSkipProbability$ of 0 produces 3TB in two days runtime.

Between time step $t_0$ and time step $t_1$, 3 047 829 new edges were added to Wikidata.
2 257 380 of these edges contain at least one node which was added to Wikidata in $t_1$.
Hence, 790 449 edges are left for which we can calculate a representation solely based on $t_0$ and therefore use for our link prediction experiment with the meta-path-based node representation.
Using the different representations, we train a logistic regression model ten times for each measurement and report the average in the following experiments.
Generally, 50\% of the new edges are used for training and 50\% are used for testing.

Our experiments are based on the argumentation of \citet{davis2011multi} that different links have different formation mechanisms.
If these formation mechanisms correlate with the different dimensions of the embedding, the model should be able to distinguish the mechanisms.
The correlation could be caused by the fact that the different dimensions represent concepts such as the domains which are determining the formation mechanisms.

\begin{figure}[!htb]
\centering
\includegraphics[scale=.7]{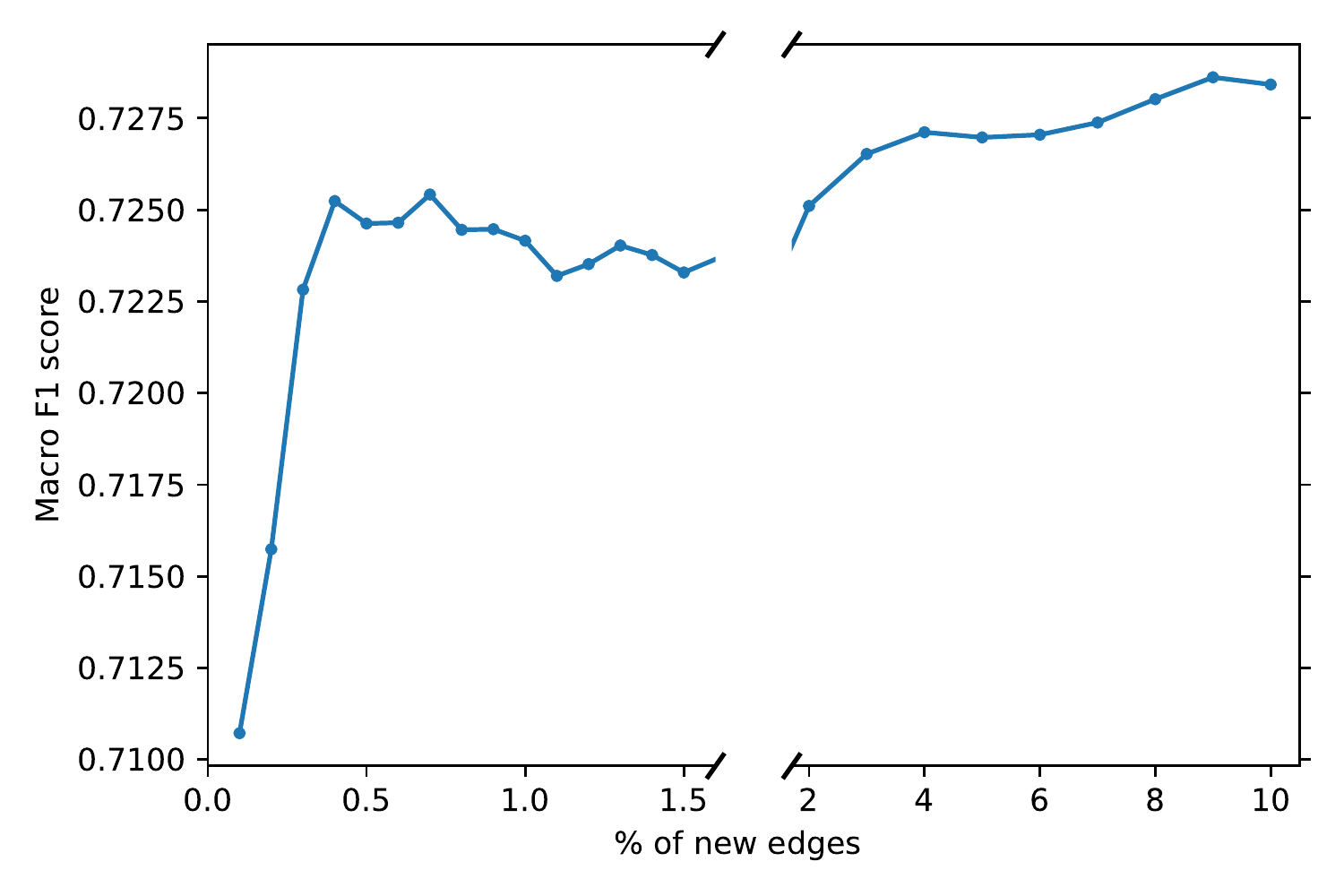}
\caption{Showing the macro F1 score for different percentages of new edges.}
\label{fig:trainsize}
\end{figure}
Our first finding is that logistic regression does not significantly profit from more than 0.5\% of the data as can be seen in Figure \ref{fig:trainsize}.
This finding allows us to perform the following experiments on a subset of all new edges.

\subsubsection{Node type-based Node Embedding}
\begin{table}
\centering
\begin{tabular}{*{6}{c}}
\toprule
emb. dim. & 16 & 32 & 64 & 128 & 256 \\
\midrule
F1 score & 0.67    & 0.67  & 0.73 & 0.72 & 0.72\\
\bottomrule
\end{tabular}
\caption{Macro F1 scores for the link prediction on 1\% of Wikidata using different embedding sizes.}
\label{tab:dim}
\end{table}
The node type-based node embedding from Section \ref{sec:nodes-types} serves as a baseline to evaluate our approach.
The experiments in Table \ref{tab:dim} show that we produce a good embedding regarding the node and edge types.
This is an indicator that we also embed the meta-paths sensibly because they are the combination of these types and define the objective function for the embedding.
Furthermore, our approach is insensitive regarding the number of dimensions and can produce a good embedding with a relatively low number of dimensions.
The best performance in both node embeddings is achieved with a dimensionality of 64.
The invariable performance could also be a sign that the logistic regression is not expressive enough for modeling more complex relations in the embeddings with a high number of dimensions.
To check this, we should repeat these experiments with a more expressive model and compare the performance.

\begin{table}
\centering
\begin{tabular}{*{5}{c}}
\toprule
Average & Concat & Hadamard & Weighted L1 & Weighted L2 \\
\midrule
0.74    & 0.740  & 0.72     & 0.48        & 0.74\\
\bottomrule
\end{tabular}
\caption{Macro F1 scores for the link prediction on 1\% of Wikidata using different vector operators to combine two node embeddings to an edge embedding.}
\label{tab:vector}
\end{table}
Additionally, the experiments in Table \ref{tab:vector} show that the embedding is insensitive with one exception to the vector operator we use to combine two node embeddings to an edge embedding.

\subsubsection{Meta-path-based Node Embedding}
\label{sec:eval-mp-node}
For the meta-path-based node embedding, we have to exclude new nodes and therefore edges involving a new node from the link prediction.
This is based on the fact that we can not mine meta-paths for these nodes in the graph at time step $t_0$.
\begin{table}
\centering
\begin{tabular}{*{5}{c}}
\toprule
emb. dim. & 128 & 256 & 512 & 1024\\
\midrule
F1 score & 0.58 & 0.57 & 0.58 & 0.58\\
\bottomrule
\end{tabular}
\caption{Macro F1 scores for the link prediction using meta-path-based node embeddings on 1\% of Wikidata with different embedding dimensions.}
\label{tab:nodes-metapaths}
\end{table}
Mining meta-paths starting in one node of the 790449 new edges with an $edgeSkipProbability$ of 0.999 yields meta-paths between 1 543 422 pairs of nodes.
If we assume that all the nodes in these edges are distinct, the mean number of meta-path sets for one node is 1.
A set of meta-paths is defined as all the meta-paths which connect two nodes.
In practice, we have 356 216 instead of the possible 1 580 898 distinct nodes. 
Therefore, we have on average 4.3 sets of meta-paths for one node.
Because we ignore the direction, we can also use the meta-paths ending in this node.

We can speed up the mining of meta-paths by introducing a stop criterion based on the number of meta-paths the algorithm has found for one node because we don't need many meta-paths to represent one node.
In the following experiment, this parameter is set to 10 meta-paths per node.
If we do not find any meta-paths for a node, we use the zero vector to represent it.
For the task of link prediction, we have to combine the embeddings of the two nodes of an edge to get the edge embedding.
With the use of the zero vector, we can not use multiplicative vector operators such as the hadamard product.
Therefore, we average the two node embeddings to derive the edge embedding.

The results in Table \ref{tab:nodes-metapaths} show that the current configuration of the node embedding based on meta-paths is significantly worse than the node type-based one.
The experiments in Section \ref{sec:eval-mp-node} show that the embeddings learn a sensible representation for at least the node and edge types.
However, the meta-path-based node embedding does not use this information for a good representation of the nodes.
The reason for this could be that the meta-paths are not adequately captured in the embedding.
This is unlikely because it captures the node and edge types despite the fact that the objective only concerns the meta-paths. 
It is more likely that we do not represent the nodes well enough in the way we combine the meta-path embedding to a node embedding.
One problem could be that  many nodes have very many meta-paths and an average of all these meta-paths is not very meaningful.
One the other side, picking some meta-paths randomly  to reduce the number probably also does not capture the node well.
One possible approach to solve this issue could be to only use the most frequent meta-paths for the node embedding.

\begin{table}
\centering
\begin{tabular}{*{6}{c}}
\toprule
emb. dim. & 16 & 32 & 64 & 128 & 256\\
\midrule
F1 score & 0.56 & 0.57 & 0.59 & 0.57 & 0.58\\
\bottomrule
\end{tabular}
\caption{Macro F1 scores for the link prediction using meta-path-based node embeddings with the minimal size of the n-grams set to 1.
The experiments were conducted on 1\% of Wikidata with different embedding dimensions.}
\label{tab:nodes-metapaths-minn}
\end{table}
When comparing the experiments in Table \ref{tab:nodes-metapaths} and \ref{tab:nodes-metapaths-minn}, we see that the inclusion of direct node and edge type embeddings neither improves nor impairs  the performance.
Therefore, it is advisable to include these embeddings to be able to calculate a node type-based node embedding using the meta-path embedding.

\subsubsection{Meta-path-based Edge Embedding}
As in Section \ref{sec:eval-mp-node}, we have to exclude new nodes from the training set and learn our prediction model on the reduced number of training samples.
With an $nodeSkipProbability$ of 0.999 resulting in 700 edges and an $edgeSkipProbability$ of 0, only for 3 edges meta-paths of length 3 are found.
We should further investigate if the meta-paths are really missing in $t_0$ or if it is a problem with the way we mine them.
The problem is most probably caused by the high probability of skipping edges.
To solve this issue, we should not search them in a breadth-first manner and instead direct the search by calculating the shortest path between these nodes.
If we make the assumption that the direction of the edges is not important as we did in our algorithm, the search can be modeled as an all-pairs-shortest-path problem on undirected and unweighted graphs.
This problem can be solved with Seidel's algorithm~\cite{seidel1995all} in $\mathcal{O}(|V|^\omega log |V|)$ where $\mathcal{O}(n^\omega)$ is the complexity of a $n \times n$ matrix multiplication.
The current upper bound of $\omega$ is roughly 2.373~\cite{le2014powers}.
When computing the shortest path, we get only one meta-path between the two nodes under the premise that the shortest path exists.
If one meta-path is not enough to represent the edge, we could formulate the search as the k shortest paths problem~\cite{eppstein1998finding}.
\citet{eppstein1998finding} proposes an algorithm for k shortest paths from one node to all others in $\mathcal{O}(|E| + |V| log|V|+ k|V|)$ . 
This algorithm can be restarted from every node to solve the all-pairs-k-shortest-path problem in $\mathcal{O}(|V|(|E| + |V| log|V|+ k|V|))=\mathcal{O}(|V||E| + |V|^2 log|V|+ k|V|^2)$ .

Further experiments should be done to determine the right size of the context window and to what extent the embedding should be regularized by excluding too frequent and infrequent meta-paths/n-grams.

\subsubsection{Comparison with Node Embedding Methods}
\begin{table}
\centering
\begin{tabular}{*{5}{c}}
\toprule
& Verse~\cite{tsitsulin2018verse} & Deepwalk~\cite{perozzi2014deepwalk} & node2vec~\cite{grover2016node2vec} & Node types \\
\midrule
1\% & 0.61    & 0.64  & - & 0.72\\
2.5\% & 0.74    & -  & - & 0.59\\
5\% & 0.83    & -  & - & 0.59\\
100\% & - & -  & - & 0.74\\
\bottomrule
\end{tabular}
\caption{Macro F1 scores for link prediction on Wikidata using differing amounts of new edges.
We compare DeepWalk, VERSE, node2vec and our method based on the node type embedding.
The vector operator is hadamard.
The node type embedding is calculated on full wikidata.
"-" marks that the method can not handle the number of nodes.}
\label{tab:wikidata-0_01}
\end{table}
The used implementations of DeepWalk\footnote{\url{https://github.com/xgfs/deepwalk-c}}, node2vec\footnote{\url{https://github.com/xgfs/node2vec-c}} and VERSE\footnote{\url{https://github.com/xgfs/verse}} can not handle the complete Wikidata dataset because the required memory exceeds the 1TB RAM of our server.
We are using the same 1\%, 2.5\% and 5\% of Wikidata for the following experiments with VERSE and DeepWalk.
As one can see in Table \ref{tab:wikidata-0_01}, DeepWalk can not handle more than 1\% and VERSE 5\% of  Wikidata.
Node2vec can not even handle 1\% of it.

The following experiments are on denser subsamples than the complete Wikidata dataset because we sample from the edges and therefore have a bias for nodes with a high degree.
We would expect a better performance of VERSE and DeepWalk on these subsamples as on the real dataset because they get more structural information.
Our method would profit from a subsampling of the nodes because a node with many neighbors also has many meta-paths which does lead to a very unspecific representation.
The increasing performance of VERSE in Table \ref{tab:wikidata-0_01} can be explained by the fact that it embeds more nodes with a similar structure and for this reason the regression model gets more training data with the same representation.
The embeddings of our method are trained on the full graph and the regression model gets only the corresponding subsets.
Our method would not profit from training it on these smaller graphs because we need multiple contexts for a meta-path to embed it sensibly.
In comparison, VERSE and DeepWalk can specialize if they are confronted with a smaller subgraph.
It is not completely clear why our method performs considerably worse on 2.5\% and 5\% in comparison with 1\% and 100\% of Wikidata.
One reason could be that we sample two subsets which are not representative for the full data or that our method can not deal with the way we subsample the graph.

\subsection{Future Experiments}
After our first experimental evaluation, further experiments should be done to compare our method with other approaches, evaluate it on other tasks, validate the generalization on other datasets and test the sensitivity of parameters we did not check.
 
 We expect that our method generalizes to other datasets well because we did not make any assumptions specific for Wikidata.
The depth of the class hierarchy is a parameter to which we would attribute a high influence on our method but did not check yet.
If the depth is too high, the embedding model does not get enough signal for infrequent node types.
If the depth is too low, only very high level classes are included in the meta-paths which are not expressive enough for the down stream task.
The $nodeSkipProbability$ and $edgeSkipProbability$ determine if the training data covers all edge and node types in the graph and if the embedding model gets a representative sample of the graph.
An open question is the influence of the amount of training data.
Additional models to compare with are the translational approaches.
Using our node embeddings, we can perform more classical node embedding experiments such as node clustering, node classification and graph reconstruction..

\subsubsection{Does the Embedding Capture the Concepts Which the Meta-paths Represent?}
With the following experiment we could check if the embedding has captured the concepts which the meta-paths represent.
The underlying assumption is that similar concepts occur together.
For example, movie-related concepts occur between two specific actors but not together with transportation-related concepts.
To verify the similarity of co-occurring concepts, we predict the presence of meta-paths instead of simple edges between two nodes.
For this experiment, we first have to mine all meta-paths between some nodes, sample one not existing meta-path for each existing one and then train a predictor.
The features for this predictor are all meta-paths between two nodes expect of one combined. 
The excluded meta-path is used as label.
If the model succeeds in predicting co-occurring concepts, we have evidence that the embedding captures the concepts.
The critical part of this experiment is that the computation of all meta-paths between some nodes is computationally very expensive.
We should first investigate if it is sufficient to probabilistically mine meta-paths between the two nodes.
The probabilistic version of the experiment is only valid if the probability of incorrectly sampling a meta-path as a negative example is low enough.

\paragraph{Correlating meta-paths}
A similar test can be performed by first searching for the most correlating meta-paths and then comparing their embeddings.
Correlation of meta-paths means that meta-paths are occurring together between pairs of nodes.
For the same reasons as in the above experiment, the embeddings should be similar.
Therefore, we can compare the quality of different embeddings by summing up a similarity measure such as cosine similarity of the embedding of the top n most correlating meta-paths.

\subsubsection{Link Type Prediction in Knowledge Graphs}
The extension of the link prediction experiment is the prediction if a link will form and which type it will have.
\citet{sun2012will} introduce this problem as relationship prediction.
This could be performed by a two stage process where we first predict the link forming and afterwards the type for the formed links.
One can reuse the predictor from Section \ref{sec:link_prediction} with this structure.
Another option would be to model "no link" as an extra edge type and then directly learn a predictor for the extended edge types.
We would expect a significantly better prediction performance of our meta-path-based approach in comparison with methods which only use the structure of the graph.
An even more complicated version of this experiment is to predict when a link will form~\cite{sun2012will}.
\pagebreak
\subsubsection{Knowledge Graph Completion}
Our technique can also be used for knowledge graph completion.
In this case, the meta-path-based edge embeddings should not suffer from the problem that nodes which are part of new edges are only sparsely connected by meta-paths.
Therefore, this task would be a good experiment to compare edge features based on node and edge embeddings.
  \section{Implementation Details}\label{sec:implementation}
In this section, we shortly describe and evaluate the pipeline of our experiments.

The graphs were imported into neo4j instances for a unified access and manipulation.
We used the \textit{wikidata-neo4j-importer}\footnote{\url{https://github.com/findie/wikidata-neo4j-importer}} to import Wikidata JSON dumps into neo4j.
This import takes up to one week and should be optimized for future experiments.

The conversion of the class hierarchy and the assignment of node types to the nodes\footnote{\url{https://github.com/Baschdl/neo4j-graph-algorithms/tree/multiTypesConversion}} as described in Section \ref{sec:evaluation} are implemented as procedures in neo4j.
The mining of meta-paths as described in Section \ref{sec:feature} with Algorithms \ref{alg:MPMining} and \ref{alg:probMPMining} is also implemented as a procedure in neo4j.
This calculation can be scaled over a cluster without synchronization overhead because each computing node can be limited to mine the meta-paths in a specific range of start nodes.
The computation of meta-paths on one computing node is even faster than the disks of the storage system can save them.

Afterwards, we combine the meta-paths between two nodes to form sentences as described in Section \ref{sec:feature} and train the embeddings using fastText\footnote{\url{https://github.com/facebookresearch/fastText}}.
Simultaneously, we search with neo4j cypher queries for new edges between $t_1$ and $t_2$ and exclude edges which new nodes\footnote{\url{https://github.com/Baschdl/metapath-embedding}}.
We perform the meta-path-based link prediction experiments using our experimentation framework\footnote{\url{https://github.com/Baschdl/bachelor-thesis-experiments/tree/metapath_embedding}}.
For a detailed description, please see~\citet{Rueckin2018Simi}.

  \section{Conclusion}\label{sec:conclusion}
In this work, we introduced the first embedding model for meta-paths.
This model is especially designed to deal with the high redundancy and high amount of meta-paths in big knowledge graphs.
Furthermore, it allows the training of machine learning models on the semantic properties of meta-paths.
Using the meta-path embedding model, we defined new node and edge type embeddings.
Additionally, we proposed new general vector representations for edges and nodes based on meta-paths.

We found problems in the first experiments regarding the presence of meta-paths between nodes which will be part of an edge in the future.
Here, we have to experiment with a meta-path mining approach using shortest path algorithms to solve these problems or prove that our assumption regarding the connectivity was wrong.
Especially our edge embeddings look promising for tasks such as link prediction.
These tasks concern edges but the current methods only use node embeddings to calculate features for them and no edge-specific features.
When dealing with real world knowledge graphs, we can not avoid to use subsampling with current methods.
But our method performs it in a way which does not hurt the applicability of our embedding in contrast to the experiments in most other works.
In our experiments, we found evidence that our method produces sensible embeddings but we have to make further experiments to work on current problems and compare it to the group of translational approaches.

Further investigation should be done concerning the influence of the data and parameter settings.
We further have to evaluate how much data our method needs and how we can adjust the collection process to it.
One could also imagine to learn a meta-path embedding by only learning node type and edge type embeddings and combining them to a meta-path embedding afterwards.
Building up on our embedding model, one could investigate if there is a better representation of parts of a meta-path than n-grams.
Another way of improving the model could be to use the structure of the class hierarchy and the property that nodes have multiple labels more directly in the embedding process.
  \section{Acknowledgements}
I thank my supervisors Davide and Prof. Emmanuel Müller for their commitment to the bachelor project and their support, our project partners for providing the initial problem and continuous support, my bachelor project team for the work together, especially Julius for jointly implementing the experiment framework and discussing specific aspects of my thesis and Freya for asking all the critical questions, and my proof readers, especially Samuel for discussing my thesis so thoroughly.

  \cleardoublepage
  \bibliography{bibliography}

\begin{thebibliography}{95}
\providecommand{\natexlab}[1]{#1}
\providecommand{\url}[1]{\texttt{#1}}
\expandafter\ifx\csname urlstyle\endcsname\relax
  \providecommand{\doi}[1]{doi: #1}\else
  \providecommand{\doi}{doi: \begingroup \urlstyle{rm}\Url}\fi

\bibitem[Adamic and Adar(2003)]{adamic2003friends}
L.~A. Adamic and E.~Adar.
\newblock Friends and neighbors on the web.
\newblock \emph{Social networks}, 25\penalty0 (3):\penalty0 211--230, 2003.

\bibitem[Ahmed et~al.(2013)Ahmed, Shervashidze, Narayanamurthy, Josifovski, and
  Smola]{ahmed2013distributed}
A.~Ahmed, N.~Shervashidze, S.~Narayanamurthy, V.~Josifovski, and A.~J. Smola.
\newblock Distributed large-scale natural graph factorization.
\newblock In \emph{Proceedings of the 22nd international conference on World
  Wide Web}, pages 37--48. ACM, 2013.

\bibitem[Akoglu et~al.(2015)Akoglu, Tong, and Koutra]{akoglu2015graph}
L.~Akoglu, H.~Tong, and D.~Koutra.
\newblock Graph based anomaly detection and description: a survey.
\newblock \emph{Data mining and knowledge discovery}, 29\penalty0 (3):\penalty0
  626--688, 2015.

\bibitem[Amaral(2008)]{amaral2008truer}
L.~A.~N. Amaral.
\newblock A truer measure of our ignorance.
\newblock \emph{Proceedings of the National Academy of Sciences}, 105\penalty0
  (19):\penalty0 6795--6796, 2008.

\bibitem[Auer et~al.(2007)Auer, Bizer, Kobilarov, Lehmann, Cyganiak, and
  Ives]{auer2007dbpedia}
S.~Auer, C.~Bizer, G.~Kobilarov, J.~Lehmann, R.~Cyganiak, and Z.~Ives.
\newblock Dbpedia: A nucleus for a web of open data.
\newblock In \emph{The semantic web}, pages 722--735. Springer, 2007.

\bibitem[Barab{\^a}si et~al.(2002)Barab{\^a}si, Jeong, N{\'e}da, Ravasz,
  Schubert, and Vicsek]{barabasi2002evolution}
A.-L. Barab{\^a}si, H.~Jeong, Z.~N{\'e}da, E.~Ravasz, A.~Schubert, and
  T.~Vicsek.
\newblock Evolution of the social network of scientific collaborations.
\newblock \emph{Physica A: Statistical mechanics and its applications},
  311\penalty0 (3-4):\penalty0 590--614, 2002.

\bibitem[Baroni and Lenci(2010)]{baroni2010distributional}
M.~Baroni and A.~Lenci.
\newblock Distributional memory: A general framework for corpus-based
  semantics.
\newblock \emph{Computational Linguistics}, 36\penalty0 (4):\penalty0 673--721,
  2010.

\bibitem[Bast et~al.(2014)Bast, B{\"a}urle, Buchhold, and
  Hau{\ss}mann]{bast2014easy}
H.~Bast, F.~B{\"a}urle, B.~Buchhold, and E.~Hau{\ss}mann.
\newblock Easy access to the freebase dataset.
\newblock In \emph{Proceedings of the 23rd International Conference on World
  Wide Web}, pages 95--98. ACM, 2014.

\bibitem[Behrens et~al.(2018)Behrens, Bischoff, Ladenburger, R{\"u}ckin,
  Seidel, Stolp, Vaichenker, Ziegler, Mottin, Aghaei,
  et~al.]{behrens2018metaexp}
F.~Behrens, S.~Bischoff, P.~Ladenburger, J.~R{\"u}ckin, L.~Seidel, F.~Stolp,
  M.~Vaichenker, A.~Ziegler, D.~Mottin, F.~Aghaei, et~al.
\newblock Metaexp: Interactive explanation and exploration of large knowledge
  graphs.
\newblock In \emph{Companion of the The Web Conference 2018 on The Web
  Conference 2018}, pages 199--202. International World Wide Web Conferences
  Steering Committee, 2018.

\bibitem[Blei et~al.(2003)Blei, Ng, and Jordan]{blei2003latent}
D.~M. Blei, A.~Y. Ng, and M.~I. Jordan.
\newblock Latent dirichlet allocation.
\newblock \emph{Journal of machine Learning research}, 3\penalty0
  (Jan):\penalty0 993--1022, 2003.

\bibitem[Blumberg and Atre(2003)]{blumberg2003problem}
R.~Blumberg and S.~Atre.
\newblock The problem with unstructured data.
\newblock \emph{Dm Review}, 13\penalty0 (42-49):\penalty0 62, 2003.

\bibitem[Bojanowski et~al.(2017)Bojanowski, Grave, Joulin, and
  Mikolov]{bojanowski2017enriching}
P.~Bojanowski, E.~Grave, A.~Joulin, and T.~Mikolov.
\newblock Enriching word vectors with subword information.
\newblock \emph{Transactions of the Association for Computational Linguistics},
  5:\penalty0 135--146, 2017.

\bibitem[Bollacker et~al.(2008)Bollacker, Evans, Paritosh, Sturge, and
  Taylor]{bollacker2008freebase}
K.~Bollacker, C.~Evans, P.~Paritosh, T.~Sturge, and J.~Taylor.
\newblock Freebase: a collaboratively created graph database for structuring
  human knowledge.
\newblock In \emph{Proceedings of the 2008 ACM SIGMOD international conference
  on Management of data}, pages 1247--1250. AcM, 2008.

\bibitem[Bordes et~al.(2011)Bordes, Weston, Collobert, Bengio,
  et~al.]{bordes2011learning}
A.~Bordes, J.~Weston, R.~Collobert, Y.~Bengio, et~al.
\newblock Learning structured embeddings of knowledge bases.
\newblock In \emph{AAAI}, volume~6, page~6, 2011.

\bibitem[Bordes et~al.(2013{\natexlab{a}})Bordes, Usunier, Garcia-Duran,
  Weston, and Yakhnenko]{bordes2013translating}
A.~Bordes, N.~Usunier, A.~Garcia-Duran, J.~Weston, and O.~Yakhnenko.
\newblock Translating embeddings for modeling multi-relational data.
\newblock In \emph{Advances in neural information processing systems}, pages
  2787--2795, 2013{\natexlab{a}}.

\bibitem[Bordes et~al.(2013{\natexlab{b}})Bordes, Usunier, Weston, and
  Yakhnenko]{Bordes2013}
A.~Bordes, N.~Usunier, J.~Weston, and O.~Yakhnenko.
\newblock {Translating Embeddings for Modeling Multi-Relational Data}.
\newblock \emph{Advances in NIPS}, 26:\penalty0 2787--2795, 2013{\natexlab{b}}.
\newblock ISSN 10495258.
\newblock \doi{10.1007/s13398-014-0173-7.2}.

\bibitem[Bordes et~al.(2014)Bordes, Glorot, Weston, and
  Bengio]{bordes2014semantic}
A.~Bordes, X.~Glorot, J.~Weston, and Y.~Bengio.
\newblock A semantic matching energy function for learning with
  multi-relational data.
\newblock \emph{Machine Learning}, 94\penalty0 (2):\penalty0 233--259, 2014.

\bibitem[Cao et~al.(2015)Cao, Lu, and Xu]{cao2015grarep}
S.~Cao, W.~Lu, and Q.~Xu.
\newblock Grarep: Learning graph representations with global structural
  information.
\newblock In \emph{Proceedings of the 24th ACM International on Conference on
  Information and Knowledge Management}, pages 891--900. ACM, 2015.

\bibitem[Davis et~al.(2011)Davis, Lichtenwalter, and Chawla]{davis2011multi}
D.~Davis, R.~Lichtenwalter, and N.~V. Chawla.
\newblock Multi-relational link prediction in heterogeneous information
  networks.
\newblock In \emph{Advances in Social Networks Analysis and Mining (ASONAM),
  2011 International Conference on}, pages 281--288. IEEE, 2011.

\bibitem[Deerwester et~al.(1990)Deerwester, Dumais, Furnas, Landauer, and
  Harshman]{deerwester1990indexing}
S.~Deerwester, S.~T. Dumais, G.~W. Furnas, T.~K. Landauer, and R.~Harshman.
\newblock Indexing by latent semantic analysis.
\newblock \emph{Journal of the American society for information science},
  41\penalty0 (6):\penalty0 391--407, 1990.

\bibitem[Dettmers et~al.(2017)Dettmers, Minervini, Stenetorp, and
  Riedel]{dettmers2017convolutional}
T.~Dettmers, P.~Minervini, P.~Stenetorp, and S.~Riedel.
\newblock Convolutional 2d knowledge graph embeddings.
\newblock \emph{arXiv preprint arXiv:1707.01476}, 2017.

\bibitem[Dong et~al.(2017)Dong, Chawla, and Swami]{dong2017metapath2vec}
Y.~Dong, N.~V. Chawla, and A.~Swami.
\newblock metapath2vec: Scalable representation learning for heterogeneous
  networks.
\newblock In \emph{Proceedings of the 23rd ACM SIGKDD International Conference
  on Knowledge Discovery and Data Mining}, pages 135--144. ACM, 2017.

\bibitem[Eppstein(1998)]{eppstein1998finding}
D.~Eppstein.
\newblock Finding the k shortest paths.
\newblock \emph{SIAM Journal on computing}, 28\penalty0 (2):\penalty0 652--673,
  1998.

\bibitem[Garcia-Duran et~al.(2015)Garcia-Duran, Bordes, Usunier, and
  Grandvalet]{Garcia-Duran2015}
A.~Garcia-Duran, A.~Bordes, N.~Usunier, and Y.~Grandvalet.
\newblock {Combining Two And Three-Way Embeddings Models for Link Prediction in
  Knowledge Bases}.
\newblock 55:\penalty0 715--742, 2015.
\newblock \doi{10.1613/jair.5013}.

\bibitem[Gardner and Mitchell(2015)]{gardner2015efficient}
M.~Gardner and T.~Mitchell.
\newblock Efficient and expressive knowledge base completion using subgraph
  feature extraction.
\newblock In \emph{Proceedings of the 2015 Conference on Empirical Methods in
  Natural Language Processing}, pages 1488--1498, 2015.

\bibitem[Graves(2013)]{graves2013generating}
A.~Graves.
\newblock Generating sequences with recurrent neural networks.
\newblock \emph{arXiv preprint arXiv:1308.0850}, 2013.

\bibitem[Grover and Leskovec(2016)]{grover2016node2vec}
A.~Grover and J.~Leskovec.
\newblock node2vec: Scalable feature learning for networks.
\newblock In \emph{Proceedings of the 22nd ACM SIGKDD international conference
  on Knowledge discovery and data mining}, pages 855--864. ACM, 2016.

\bibitem[Guo et~al.(2015)Guo, Wang, Wang, Wang, and Guo]{guo2015semantically}
S.~Guo, Q.~Wang, B.~Wang, L.~Wang, and L.~Guo.
\newblock Semantically smooth knowledge graph embedding.
\newblock In \emph{Proceedings of the 53rd Annual Meeting of the Association
  for Computational Linguistics and the 7th International Joint Conference on
  Natural Language Processing (Volume 1: Long Papers)}, volume~1, pages 84--94,
  2015.

\bibitem[Gupta et~al.(2016)Gupta, Varma, et~al.]{gupta2016doc2sent2vec}
M.~Gupta, V.~Varma, et~al.
\newblock Doc2sent2vec: A novel two-phase approach for learning document
  representation.
\newblock In \emph{Proceedings of the 39th International ACM SIGIR conference
  on Research and Development in Information Retrieval}, pages 809--812. ACM,
  2016.

\bibitem[Guu et~al.(2015)Guu, Miller, and Liang]{guu2015traversing}
K.~Guu, J.~Miller, and P.~Liang.
\newblock Traversing knowledge graphs in vector space.
\newblock In \emph{Proceedings of the 2015 Conference on Empirical Methods in
  Natural Language Processing}, pages 318--327, 2015.

\bibitem[Harris(1954)]{harris1954distributional}
Z.~S. Harris.
\newblock Distributional structure.
\newblock \emph{Word}, 10\penalty0 (2-3):\penalty0 146--162, 1954.

\bibitem[Hoffart et~al.(2013)Hoffart, Suchanek, Berberich, and
  Weikum]{hoffart2013yago2}
J.~Hoffart, F.~M. Suchanek, K.~Berberich, and G.~Weikum.
\newblock Yago2: A spatially and temporally enhanced knowledge base from
  wikipedia.
\newblock \emph{Artificial Intelligence}, 194:\penalty0 28--61, 2013.

\bibitem[Hofmann(1999)]{hofmann1999probabilistic}
T.~Hofmann.
\newblock Probabilistic latent semantic indexing.
\newblock In \emph{Proceedings of the 22nd annual international ACM SIGIR
  conference on Research and development in information retrieval}, pages
  50--57. ACM, 1999.

\bibitem[Huang et~al.(2016)Huang, Zheng, Cheng, Sun, Mamoulis, and
  Li]{huang2016meta}
Z.~Huang, Y.~Zheng, R.~Cheng, Y.~Sun, N.~Mamoulis, and X.~Li.
\newblock Meta structure: Computing relevance in large heterogeneous
  information networks.
\newblock In \emph{Proceedings of the 22nd ACM SIGKDD International Conference
  on Knowledge Discovery and Data Mining}, pages 1595--1604. ACM, 2016.

\bibitem[Katz(1953)]{katz1953new}
L.~Katz.
\newblock A new status index derived from sociometric analysis.
\newblock \emph{Psychometrika}, 18\penalty0 (1):\penalty0 39--43, 1953.

\bibitem[Kim et~al.(2016)Kim, Jernite, Sontag, and Rush]{kim2016character}
Y.~Kim, Y.~Jernite, D.~Sontag, and A.~M. Rush.
\newblock Character-aware neural language models.
\newblock In \emph{AAAI}, pages 2741--2749, 2016.

\bibitem[Kong et~al.(2012)Kong, Yu, Ding, and Wild]{kong2012meta}
X.~Kong, P.~S. Yu, Y.~Ding, and D.~J. Wild.
\newblock Meta path-based collective classification in heterogeneous
  information networks.
\newblock In \emph{Proceedings of the 21st ACM international conference on
  Information and knowledge management}, pages 1567--1571. ACM, 2012.

\bibitem[Lao and Cohen(2010)]{lao2010relational}
N.~Lao and W.~W. Cohen.
\newblock Relational retrieval using a combination of path-constrained random
  walks.
\newblock \emph{Machine learning}, 81\penalty0 (1):\penalty0 53--67, 2010.

\bibitem[Le and Mikolov(2014)]{le2014distributed}
Q.~Le and T.~Mikolov.
\newblock Distributed representations of sentences and documents.
\newblock In \emph{International Conference on Machine Learning}, pages
  1188--1196, 2014.

\bibitem[Le~Gall(2014)]{le2014powers}
F.~Le~Gall.
\newblock Powers of tensors and fast matrix multiplication.
\newblock In \emph{Proceedings of the 39th international symposium on symbolic
  and algebraic computation}, pages 296--303. ACM, 2014.

\bibitem[Lenat and Feigenbaum(1992)]{lenat1992thresholds}
D.~Lenat and E.~Feigenbaum.
\newblock On the thresholds of knowledge.
\newblock \emph{Foundations of Artificial Intelligence, MIT Press, Cambridge,
  MA}, pages 185--250, 1992.

\bibitem[Levy and Goldberg(2014)]{levy2014neural}
O.~Levy and Y.~Goldberg.
\newblock Neural word embedding as implicit matrix factorization.
\newblock In \emph{Advances in neural information processing systems}, pages
  2177--2185, 2014.

\bibitem[Liben-Nowell and Kleinberg(2007)]{liben2007link}
D.~Liben-Nowell and J.~Kleinberg.
\newblock The link-prediction problem for social networks.
\newblock \emph{Journal of the American society for information science and
  technology}, 58\penalty0 (7):\penalty0 1019--1031, 2007.

\bibitem[Lichtenwalter et~al.(2010)Lichtenwalter, Lussier, and
  Chawla]{lichtenwalter2010new}
R.~N. Lichtenwalter, J.~T. Lussier, and N.~V. Chawla.
\newblock New perspectives and methods in link prediction.
\newblock In \emph{Proceedings of the 16th ACM SIGKDD international conference
  on Knowledge discovery and data mining}, pages 243--252. ACM, 2010.

\bibitem[Lin et~al.(2017)Lin, Liu, Wang, Yue, and Lin]{Lin2017}
H.~Lin, Y.~Liu, W.~Wang, Y.~Yue, and Z.~Lin.
\newblock {Learning Entity and Relation Embeddings for Knowledge Resolution}.
\newblock \emph{Procedia Computer Science}, 108:\penalty0 345--354, 2017.
\newblock ISSN 18770509.
\newblock \doi{10.1016/j.procs.2017.05.045}.

\bibitem[Lin et~al.(2015)Lin, Liu, Luan, Sun, Rao, and Liu]{lin2015modeling}
Y.~Lin, Z.~Liu, H.~Luan, M.~Sun, S.~Rao, and S.~Liu.
\newblock Modeling relation paths for representation learning of knowledge
  bases.
\newblock In \emph{Proceedings of the 2015 Conference on Empirical Methods in
  Natural Language Processing}, pages 705--714, 2015.

\bibitem[Liu et~al.(2014)Liu, Yu, Guo, and Sun]{liu2014meta}
X.~Liu, Y.~Yu, C.~Guo, and Y.~Sun.
\newblock Meta-path-based ranking with pseudo relevance feedback on
  heterogeneous graph for citation recommendation.
\newblock In \emph{Proceedings of the 23rd acm international conference on
  conference on information and knowledge management}, pages 121--130. ACM,
  2014.

\bibitem[L{\"u} and Zhou(2011)]{lu2011link}
L.~L{\"u} and T.~Zhou.
\newblock Link prediction in complex networks: A survey.
\newblock \emph{Physica A: statistical mechanics and its applications},
  390\penalty0 (6):\penalty0 1150--1170, 2011.

\bibitem[Luo et~al.(2015)Luo, Wang, Wang, and Guo]{luo2015context}
Y.~Luo, Q.~Wang, B.~Wang, and L.~Guo.
\newblock Context-dependent knowledge graph embedding.
\newblock In \emph{Proceedings of the 2015 Conference on Empirical Methods in
  Natural Language Processing}, pages 1656--1661, 2015.

\bibitem[Meng et~al.(2015)Meng, Cheng, Maniu, Senellart, and
  Zhang]{meng2015discovering}
C.~Meng, R.~Cheng, S.~Maniu, P.~Senellart, and W.~Zhang.
\newblock Discovering meta-paths in large heterogeneous information networks.
\newblock In \emph{WWW}, pages 754--764, 2015.

\bibitem[Mikolov et~al.()Mikolov, Sutskever, Deoras, Le, and
  Kombrink]{mikolov2012subword}
T.~Mikolov, I.~Sutskever, A.~Deoras, H.-S. Le, and S.~Kombrink.
\newblock Subword language modeling with neural networks.

\bibitem[Mikolov et~al.(2013{\natexlab{a}})Mikolov, Chen, Corrado, and
  Dean]{mikolov2013efficient}
T.~Mikolov, K.~Chen, G.~Corrado, and J.~Dean.
\newblock Efficient estimation of word representations in vector space.
\newblock \emph{arXiv preprint arXiv:1301.3781}, 2013{\natexlab{a}}.

\bibitem[Mikolov et~al.(2013{\natexlab{b}})Mikolov, Sutskever, Chen, Corrado,
  and Dean]{mikolov2013distributed}
T.~Mikolov, I.~Sutskever, K.~Chen, G.~S. Corrado, and J.~Dean.
\newblock Distributed representations of words and phrases and their
  compositionality.
\newblock In \emph{Advances in neural information processing systems}, pages
  3111--3119, 2013{\natexlab{b}}.

\bibitem[Murphy et~al.(2012)Murphy, Talukdar, and Mitchell]{murphy2012learning}
B.~Murphy, P.~Talukdar, and T.~Mitchell.
\newblock Learning effective and interpretable semantic models using
  non-negative sparse embedding.
\newblock \emph{Proceedings of COLING 2012}, pages 1933--1950, 2012.

\bibitem[Neelakantan et~al.(2015)Neelakantan, Roth, and
  McCallum]{neelakantan2015compositional}
A.~Neelakantan, B.~Roth, and A.~McCallum.
\newblock Compositional vector space models for knowledge base completion.
\newblock In \emph{Proceedings of the 53rd Annual Meeting of the Association
  for Computational Linguistics and the 7th International Joint Conference on
  Natural Language Processing (Volume 1: Long Papers)}, volume~1, pages
  156--166, 2015.

\bibitem[Newman(2001)]{newman2001clustering}
M.~E. Newman.
\newblock Clustering and preferential attachment in growing networks.
\newblock \emph{Physical review E}, 64\penalty0 (2):\penalty0 025102, 2001.

\bibitem[Nguyen(2017)]{nguyen2017overview}
D.~Q. Nguyen.
\newblock An overview of embedding models of entities and relationships for
  knowledge base completion.
\newblock \emph{arXiv preprint arXiv:1703.08098}, 2017.

\bibitem[Nguyen et~al.(2016)Nguyen, Sirts, Qu, and
  Johnson]{nguyen2016neighborhood}
D.~Q. Nguyen, K.~Sirts, L.~Qu, and M.~Johnson.
\newblock Neighborhood mixture model for knowledge base completion.
\newblock \emph{CoNLL 2016}, page~40, 2016.

\bibitem[Nguyen et~al.(2018)Nguyen, Nguyen, Nguyen, and Phung]{nguyen2018novel}
D.~Q. Nguyen, T.~D. Nguyen, D.~Q. Nguyen, and D.~Phung.
\newblock A novel embedding model for knowledge base completion based on
  convolutional neural network.
\newblock In \emph{Proceedings of the 2018 Conference of the North American
  Chapter of the Association for Computational Linguistics: Human Language
  Technologies, Volume 2 (Short Papers)}, volume~2, pages 327--333, 2018.

\bibitem[Nickel et~al.(2016{\natexlab{a}})Nickel, Murphy, Tresp, and
  Gabrilovich]{nickel2016review}
M.~Nickel, K.~Murphy, V.~Tresp, and E.~Gabrilovich.
\newblock A review of relational machine learning for knowledge graphs.
\newblock \emph{Proceedings of the IEEE}, 104\penalty0 (1):\penalty0 11--33,
  2016{\natexlab{a}}.

\bibitem[Nickel et~al.(2016{\natexlab{b}})Nickel, Rosasco, Poggio,
  et~al.]{nickel2016holographic}
M.~Nickel, L.~Rosasco, T.~A. Poggio, et~al.
\newblock Holographic embeddings of knowledge graphs.
\newblock In \emph{AAAI}, volume~2, pages 3--2, 2016{\natexlab{b}}.

\bibitem[Ou et~al.(2016)Ou, Cui, Pei, Zhang, and Zhu]{ou2016asymmetric}
M.~Ou, P.~Cui, J.~Pei, Z.~Zhang, and W.~Zhu.
\newblock Asymmetric transitivity preserving graph embedding.
\newblock In \emph{Proceedings of the 22nd ACM SIGKDD international conference
  on Knowledge discovery and data mining}, pages 1105--1114. ACM, 2016.

\bibitem[Pennington et~al.(2014)Pennington, Socher, and
  Manning]{pennington2014glove}
J.~Pennington, R.~Socher, and C.~Manning.
\newblock Glove: Global vectors for word representation.
\newblock In \emph{Proceedings of the 2014 conference on empirical methods in
  natural language processing (EMNLP)}, pages 1532--1543, 2014.

\bibitem[Perozzi et~al.(2014)Perozzi, Al-Rfou, and Skiena]{perozzi2014deepwalk}
B.~Perozzi, R.~Al-Rfou, and S.~Skiena.
\newblock Deepwalk: Online learning of social representations.
\newblock In \emph{Proceedings of the 20th ACM SIGKDD international conference
  on Knowledge discovery and data mining}, pages 701--710. ACM, 2014.

\bibitem[Rückin(2018)]{Rueckin2018Simi}
J.~Rückin.
\newblock Similarity explanation and exploration in a heterogeneous information
  network, 2018.

\bibitem[Salle et~al.(2016)Salle, Idiart, and Villavicencio]{salle2016matrix}
A.~Salle, M.~Idiart, and A.~Villavicencio.
\newblock Matrix factorization using window sampling and negative sampling for
  improved word representations.
\newblock \emph{arXiv preprint arXiv:1606.00819}, 2016.

\bibitem[Seidel(1995)]{seidel1995all}
R.~Seidel.
\newblock On the all-pairs-shortest-path problem in unweighted undirected
  graphs.
\newblock \emph{Journal of computer and system sciences}, 51\penalty0
  (3):\penalty0 400--403, 1995.

\bibitem[Shi et~al.(2012)Shi, Kong, Yu, Xie, and Wu]{shi2012relevance}
C.~Shi, X.~Kong, P.~S. Yu, S.~Xie, and B.~Wu.
\newblock Relevance search in heterogeneous networks.
\newblock In \emph{Proceedings of the 15th International Conference on
  Extending Database Technology}, pages 180--191. ACM, 2012.

\bibitem[Shi et~al.(2014)Shi, Kong, Huang, Philip, and Wu]{shi2014hetesim}
C.~Shi, X.~Kong, Y.~Huang, S.~Y. Philip, and B.~Wu.
\newblock Hetesim: A general framework for relevance measure in heterogeneous
  networks.
\newblock \emph{IEEE Trans. Knowl. Data Eng.}, 26\penalty0 (10):\penalty0
  2479--2492, 2014.

\bibitem[Shi et~al.(2017)Shi, Li, Zhang, Sun, and Philip]{shi2017survey}
C.~Shi, Y.~Li, J.~Zhang, Y.~Sun, and S.~Y. Philip.
\newblock A survey of heterogeneous information network analysis.
\newblock \emph{IEEE Transactions on Knowledge and Data Engineering},
  29\penalty0 (1):\penalty0 17--37, 2017.

\bibitem[Socher et~al.(2013)Socher, Chen, Manning, and Ng]{socher2013reasoning}
R.~Socher, D.~Chen, C.~D. Manning, and A.~Ng.
\newblock Reasoning with neural tensor networks for knowledge base completion.
\newblock In \emph{Advances in neural information processing systems}, pages
  926--934, 2013.

\bibitem[Stumpf et~al.(2008)Stumpf, Thorne, de~Silva, Stewart, An, Lappe, and
  Wiuf]{stumpf2008estimating}
M.~P. Stumpf, T.~Thorne, E.~de~Silva, R.~Stewart, H.~J. An, M.~Lappe, and
  C.~Wiuf.
\newblock Estimating the size of the human interactome.
\newblock \emph{Proceedings of the National Academy of Sciences}, 105\penalty0
  (19):\penalty0 6959--6964, 2008.

\bibitem[Suchanek et~al.(2007)Suchanek, Kasneci, and Weikum]{suchanek2007yago}
F.~M. Suchanek, G.~Kasneci, and G.~Weikum.
\newblock Yago: a core of semantic knowledge.
\newblock In \emph{Proceedings of the 16th international conference on World
  Wide Web}, pages 697--706. ACM, 2007.

\bibitem[Sun and Han(2012)]{sun2012mining}
Y.~Sun and J.~Han.
\newblock Mining heterogeneous information networks: principles and
  methodologies.
\newblock \emph{Synthesis Lectures on Data Mining and Knowledge Discovery},
  3\penalty0 (2):\penalty0 1--159, 2012.

\bibitem[Sun and Han(2013)]{sun2013mining}
Y.~Sun and J.~Han.
\newblock Mining heterogeneous information networks: a structural analysis
  approach.
\newblock \emph{ACM SIGKDD Explorations Newsletter}, 14\penalty0 (2):\penalty0
  20--28, 2013.

\bibitem[Sun et~al.(2009)Sun, Yu, and Han]{sun2009ranking}
Y.~Sun, Y.~Yu, and J.~Han.
\newblock Ranking-based clustering of heterogeneous information networks with
  star network schema.
\newblock In \emph{Proceedings of the 15th ACM SIGKDD international conference
  on Knowledge discovery and data mining}, pages 797--806. ACM, 2009.

\bibitem[Sun et~al.(2011{\natexlab{a}})Sun, Barber, Gupta, Aggarwal, and
  Han]{sun2011co}
Y.~Sun, R.~Barber, M.~Gupta, C.~C. Aggarwal, and J.~Han.
\newblock Co-author relationship prediction in heterogeneous bibliographic
  networks.
\newblock In \emph{Advances in Social Networks Analysis and Mining (ASONAM),
  2011 International Conference on}, pages 121--128. IEEE, 2011{\natexlab{a}}.

\bibitem[Sun et~al.(2011{\natexlab{b}})Sun, Han, Yan, Yu, and
  Wu]{sun2011pathsim}
Y.~Sun, J.~Han, X.~Yan, P.~S. Yu, and T.~Wu.
\newblock Pathsim: Meta path-based top-k similarity search in heterogeneous
  information networks.
\newblock \emph{PVLDB}, 4\penalty0 (11):\penalty0 992--1003,
  2011{\natexlab{b}}.

\bibitem[Sun et~al.(2012)Sun, Han, Aggarwal, and Chawla]{sun2012will}
Y.~Sun, J.~Han, C.~C. Aggarwal, and N.~V. Chawla.
\newblock When will it happen?: relationship prediction in heterogeneous
  information networks.
\newblock In \emph{Proceedings of the fifth ACM international conference on Web
  search and data mining}, pages 663--672. ACM, 2012.

\bibitem[Sun et~al.(2013)Sun, Norick, Han, Yan, Yu, and Yu]{sun2013pathselclus}
Y.~Sun, B.~Norick, J.~Han, X.~Yan, P.~S. Yu, and X.~Yu.
\newblock Pathselclus: Integrating meta-path selection with user-guided object
  clustering in heterogeneous information networks.
\newblock \emph{ACM Transactions on Knowledge Discovery from Data (TKDD)},
  7\penalty0 (3):\penalty0 11, 2013.

\bibitem[Sutskever et~al.(2011)Sutskever, Martens, and
  Hinton]{sutskever2011generating}
I.~Sutskever, J.~Martens, and G.~E. Hinton.
\newblock Generating text with recurrent neural networks.
\newblock In \emph{Proceedings of the 28th International Conference on Machine
  Learning (ICML-11)}, pages 1017--1024, 2011.

\bibitem[Tang et~al.(2015)Tang, Qu, Wang, Zhang, Yan, and Mei]{tang2015line}
J.~Tang, M.~Qu, M.~Wang, M.~Zhang, J.~Yan, and Q.~Mei.
\newblock Line: Large-scale information network embedding.
\newblock In \emph{Proceedings of the 24th International Conference on World
  Wide Web}, pages 1067--1077. International World Wide Web Conferences
  Steering Committee, 2015.

\bibitem[Toutanova and Chen(2015)]{toutanova2015observed}
K.~Toutanova and D.~Chen.
\newblock Observed versus latent features for knowledge base and text
  inference.
\newblock In \emph{Proceedings of the 3rd Workshop on Continuous Vector Space
  Models and their Compositionality}, pages 57--66, 2015.

\bibitem[Toutanova et~al.(2016)Toutanova, Lin, Yih, Poon, and
  Quirk]{toutanova2016compositional}
K.~Toutanova, V.~Lin, W.-t. Yih, H.~Poon, and C.~Quirk.
\newblock Compositional learning of embeddings for relation paths in knowledge
  base and text.
\newblock In \emph{Proceedings of the 54th Annual Meeting of the Association
  for Computational Linguistics (Volume 1: Long Papers)}, volume~1, pages
  1434--1444, 2016.

\bibitem[Tsitsulin et~al.(2018)Tsitsulin, Mottin, Karras, and
  M{\"u}ller]{tsitsulin2018verse}
A.~Tsitsulin, D.~Mottin, P.~Karras, and E.~M{\"u}ller.
\newblock Verse: Versatile graph embeddings from similarity measures.
\newblock In \emph{Proceedings of the 2018 World Wide Web Conference on World
  Wide Web}, pages 539--548. International World Wide Web Conferences Steering
  Committee, 2018.

\bibitem[Turney and Pantel(2010)]{turney2010frequency}
P.~D. Turney and P.~Pantel.
\newblock From frequency to meaning: Vector space models of semantics.
\newblock \emph{Journal of artificial intelligence research}, 37:\penalty0
  141--188, 2010.

\bibitem[Vrande{\v{c}}i{\'c} and Kr{\"o}tzsch(2014)]{vrandevcic2014wikidata}
D.~Vrande{\v{c}}i{\'c} and M.~Kr{\"o}tzsch.
\newblock Wikidata: a free collaborative knowledgebase.
\newblock \emph{Communications of the ACM}, 57\penalty0 (10):\penalty0 78--85,
  2014.

\bibitem[Wang et~al.(2016)Wang, Liu, Luo, Wang, and Lin]{wang2016knowledge}
Q.~Wang, J.~Liu, Y.~Luo, B.~Wang, and C.-Y. Lin.
\newblock Knowledge base completion via coupled path ranking.
\newblock In \emph{Proceedings of the 54th Annual Meeting of the Association
  for Computational Linguistics (Volume 1: Long Papers)}, volume~1, pages
  1308--1318, 2016.

\bibitem[Wang et~al.(2017)Wang, Mao, Wang, and Guo]{wang2017knowledge}
Q.~Wang, Z.~Mao, B.~Wang, and L.~Guo.
\newblock Knowledge graph embedding: A survey of approaches and applications.
\newblock \emph{IEEE Transactions on Knowledge and Data Engineering},
  29\penalty0 (12):\penalty0 2724--2743, 2017.

\bibitem[Wang et~al.(2014)Wang, Zhang, Feng, and Chen]{Wang2014}
Z.~Wang, J.~Zhang, J.~Feng, and Z.~Chen.
\newblock {Knowledge Graph Embedding by Translating on Hyperplanes}.
\newblock \emph{AAAI Conference on Artificial Intelligence}, pages 1112--1119,
  2014.

\bibitem[Xie et~al.(2016)Xie, Liu, and Sun]{xie2016representation}
R.~Xie, Z.~Liu, and M.~Sun.
\newblock Representation learning of knowledge graphs with hierarchical types.
\newblock In \emph{IJCAI}, pages 2965--2971, 2016.

\bibitem[Yu et~al.(2012{\natexlab{a}})Yu, Gu, Zhou, and Han]{yu2012citation}
X.~Yu, Q.~Gu, M.~Zhou, and J.~Han.
\newblock Citation prediction in heterogeneous bibliographic networks.
\newblock In \emph{Proceedings of the 2012 SIAM International Conference on
  Data Mining}, pages 1119--1130. SIAM, 2012{\natexlab{a}}.

\bibitem[Yu et~al.(2012{\natexlab{b}})Yu, Sun, Norick, Mao, and
  Han]{yu2012user}
X.~Yu, Y.~Sun, B.~Norick, T.~Mao, and J.~Han.
\newblock User guided entity similarity search using meta-path selection in
  heterogeneous information networks.
\newblock In \emph{Proceedings of the 21st ACM international conference on
  Information and knowledge management}, pages 2025--2029. Acm,
  2012{\natexlab{b}}.

\bibitem[Zhang et~al.(2014)Zhang, Yu, and Zhou]{zhang2014meta}
J.~Zhang, P.~S. Yu, and Z.-H. Zhou.
\newblock Meta-path based multi-network collective link prediction.
\newblock In \emph{Proceedings of the 20th ACM SIGKDD international conference
  on Knowledge discovery and data mining}, pages 1286--1295. ACM, 2014.

\bibitem[Ziegler(2018)]{Ziegler2018Active}
A.~Ziegler.
\newblock Active learning with curriculum on knowledge graphs, 2018.

\end{thebibliography}
  \cleardoublepage
  \thispagestyle{empty-page}
  \section*{German abstract}

\begin{otherlanguage}{german}
In dieser Arbeit untersuchen wir das Lernen von Merkmalen auf heterogenen Wissensgraphen.
Diese Merkmale können für Aufgaben wie die Vorhersage von Verbindungen, Klassifikation und Clustering verwendet werden.
Wissensgraphen bieten eine reichhaltige Semantik, die in den Kanten- und Knotentypen codiert ist.
Meta-Pfade bestehen aus diesen Typen und stellen eine Abstraktion von Pfaden im Graphen dar.

Bisher können Meta-Pfade nur als kategorische Merkmale mit hoher Redundanz verwendet werden.
Wir stellen einen Ansatz zum Lernen von semantischen Repräsentationen von Meta-Pfaden vor.
Aktuelle Methoden können nur Repräsentation für Knoten und Kantentypen lernen.
Unser Ansatz benutzt ein erweitertes Skipgram-Modell um, trotz hoher Redundanz und Anzahl, Merkmale für Meta-Pfade zu lernen.
Wir evaluieren unserer Methode mit der Vorhersage von Verbindungen auf Wikidata.
Diese Experimente zeigen, dass wir eine sinnvolle Repräsentation lernen.
Allerdings müssen wir die Merkmale für die Vorhersage von Verbindungen verbessern und weitere Experimente durchführen.
\end{otherlanguage}
  \cleardoublepage
  \thispagestyle{empty-page}
  \section*{Selbstständigkeitserklärung}

\begin{otherlanguage}{german}
Hiermit erkläre ich, die vorliegende Arbeit
selbstständig angefertigt, nicht anderweitig zu Prüfungszwecken vorgelegt und
keine anderen als die angegebenen Hilfsmittel verwendet zu haben. Sämtliche
wissentlich verwendete Textausschnitte, Zitate oder Inhalte anderer Verfasser
wurden ausdrücklich als solche gekennzeichnet.
\end{otherlanguage}

Potsdam, den 30. Juli 2018\\[2cm] 
\begin{tabular}{p{5cm}}\hline
\centering\footnotesize Sebastian Bischoff
\end{tabular}

\end{document}